\relax
\documentclass[onecolumn]{article} 
\usepackage{times}  
\usepackage{helvet} 
\usepackage{courier}  
\usepackage[hyphens]{url}  
\usepackage{graphicx} 
\usepackage{graphicx}  
\usepackage{caption} 
\usepackage[margin=2.8cm]{geometry}
\usepackage{cite}
\usepackage{hyperref}

\usepackage{latexsym}
\usepackage{amscd}
\usepackage{amsmath}
\usepackage{amssymb}
\usepackage{amsthm}
\usepackage{mathrsfs}
\usepackage[ruled,vlined]{algorithm2e}
\usepackage{tikz}
\usepackage{authblk}

\usepackage[switch]{lineno}
\newcommand*\patchAmsMathEnvironmentForLineno[1]{
  \expandafter\let\csname old#1\expandafter\endcsname\csname #1\endcsname
  \expandafter\let\csname oldend#1\expandafter\endcsname\csname end#1\endcsname
  \renewenvironment{#1}
     {\linenomath\csname old#1\endcsname}
     {\csname oldend#1\endcsname\endlinenomath}}
\newcommand*\patchBothAmsMathEnvironmentsForLineno[1]{
  \patchAmsMathEnvironmentForLineno{#1}
  \patchAmsMathEnvironmentForLineno{#1*}}
\AtBeginDocument{
\patchBothAmsMathEnvironmentsForLineno{equation}
\patchBothAmsMathEnvironmentsForLineno{align}
\patchBothAmsMathEnvironmentsForLineno{flalign}
\patchBothAmsMathEnvironmentsForLineno{alignat}
\patchBothAmsMathEnvironmentsForLineno{gather}
\patchBothAmsMathEnvironmentsForLineno{multline}
}

\newcommand{\set}[1]{\{ #1 \}}
\newcommand{\ind}[1]{\mathbb{I}\left[ #1 \right]}
\newcommand{\supp}[1]{\mathrm{supp}( #1 )}
\newcommand{\hx}{\hat{x}}
\DeclareMathOperator*{\argmin}{\operatorname{arg\, min}}
\DeclareMathOperator*{\minimize}{\operatorname{minimize}}

\newcommand{\bbR}{\mathbb{R}}
\newcommand{\myendbox}{\hfill \ensuremath{\Box}}

\theoremstyle{definition}
\newtheorem{problem}{Problem}

\newtheorem{example}{Example}

\usepackage{cleveref}
\crefname{algorithm}{Algorithm}{Algorithms}
\crefname{table}{Table}{Tables}
\crefname{figure}{Figure}{Figures}

\usepackage{prettyref}
\newcommand{\pref}{\prettyref}
\newrefformat{const}{Constraint~\eqref{#1}}

\usepackage{subfig}
\usepackage{multirow}
\usepackage{booktabs}

\renewcommand{\epsilon}{\varepsilon}

\title{Ordered Counterfactual Explanation by \\ Mixed-Integer Linear Optimization}

\author[1]{Kentaro Kanamori\footnote{e-mail: \texttt{\{kanamori, arim\}@ist.hokudai.ac.jp}}}
\author[2]{Takuya Takagi\footnote{e-mail: \texttt{\{takagi.takuya, ken-kobayashi, ike.yuichi, uemura.kento\}@fujitsu.com}}}
\author[2,3]{Ken Kobayashi${}^\ddagger$}
\author[2]{\\Yuichi Ike${}^\ddagger$}
\author[2]{Kento Uemura${}^\ddagger$}
\author[1]{Hiroki Arimura${}^{\dagger}$}

\affil[1]{Hokkaido University}
\affil[2]{Fujitsu Laboratories Ltd.}
\affil[3]{Tokyo Institute of Technology}

\date{}

\begin{document}
\maketitle

\begin{abstract}
Post-hoc explanation methods for machine learning models have been widely used to support decision-making. 
One of the popular methods is \emph{Counterfactual Explanation (CE)}, also known as \emph{Actionable Recourse}, which provides a user with a perturbation vector of features that alters the prediction result. 
Given a perturbation vector, a user can interpret it as an ``\emph{action}" for obtaining one's desired decision result. 
In practice, however, showing only a perturbation vector is often insufficient for users to execute the action. 
The reason is that if there is an asymmetric interaction among features, such as causality, the total cost of the action is expected to depend on the order of changing features. 
Therefore, practical CE methods are required to provide an appropriate order of changing features in addition to a perturbation vector. 
For this purpose, we propose a new framework called \emph{Ordered Counterfactual Explanation (OrdCE)}. 
We introduce a new objective function that evaluates a pair of an action and an order based on feature interaction. 
To extract an optimal pair, we propose a mixed-integer linear optimization approach with our objective function. 
Numerical experiments on real datasets demonstrated the effectiveness of our OrdCE in comparison with unordered CE methods. 
\end{abstract}

\section{Introduction}\label{sec:intro}

Complex machine learning models such as neural networks and tree ensembles are widely used in critical decision-making tasks (e.g., medical diagnosis and loan approval). 
Thus, post-hoc methods for extracting explanations from an individual prediction of these models have been attracting much attention for the last few years~\cite{Ribeiro:KDD2016,Lundberg:NIPS2017,Koh:ICML2017,Ribeiro:AAAI2018}.
To provide a user with a better insight into future improvement, a post-hoc method needs to show not only why undesirable predictions are given, but also how to act to obtain a desired prediction result~\cite{Doshi-Velez:arxiv2017,Miller:AI2019}.

\begin{figure}[t]
    \centering

    \subfloat[Features]{
        \includegraphics[width=0.3\linewidth]{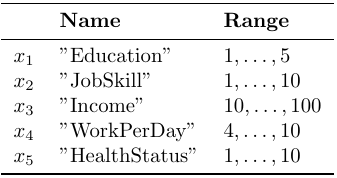}
    }
    \subfloat[Causal DAG]{
        \begin{tikzpicture}
        \node[draw, rectangle,rounded corners=3pt] (a) at (0,1.0) {\small Education};
        \node[draw, rectangle,rounded corners=3pt] (b) at (0,0) {\small JobSkill};
        \node[draw, rectangle,rounded corners=3pt] (c) at (0,-1.0) {\small Income};
        \node[draw, rectangle,rounded corners=3pt] (d) at (2.,0) {\small WorkPerDay};
        \node[draw, rectangle,rounded corners=3pt] (e) at (2.,-1.0) {\small HealthStatus};
        \draw[very thick, ->] (a)--(b) node [midway, left] {\footnotesize $1.00$};
        \draw[very thick, ->] (b)--(c) node [midway, left] {\footnotesize $6.00$};
        \draw[very thick, ->] (d)--(c) node [midway, right] {\footnotesize \ $4.00$};
        \draw[very thick, ->] (d)--(e) node [midway, right] {\footnotesize $-0.50$};
    \end{tikzpicture}
    }
    \caption{Features and the causal DAG of our synthetic credit loan approval dataset.
            The task is to predict whether an individual's credit loan will be approved.
            We labeled each individual depending on one's values of ``Income" and ``HealthStatus".
            }
    \label{fig:demo}
\end{figure}
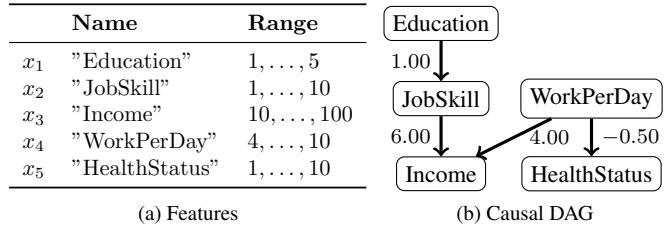

One of the post-hoc methods that show an action to obtain
a desired outcome is \emph{Counterfactual Explanation~(CE)}~\cite{Wachter:HJLT2018}, also known as \emph{Actionable Recourse}~\cite{Ustun:FAT*2019}. 
Consider an example of a synthetic credit loan approval dataset shown in~\cref{fig:demo}. 
Imagine a situation that a user receives an undesired prediction $H(\hx) \not= y^\star$ for a target label~$y^\star$ from a trained model~$H$, which means denial of credit loan, on an instance $\hx$ related to one's current livelihood. 
We want to provide the user with advice $a$ on changes of features such as ``Income" and ``JobSkill" so that the user can change one's current status $\hx$ to obtain the desired outcome $H(\hx+a) = y^\star$~\cite{Ustun:FAT*2019}.

To achieve this goal, most of the existing CE methods find a perturbation vector $a^\star \in \mathcal{A}$, called an \emph{action}, as an optimal solution of the following optimization problem:
\begin{align*}
        a^\star := \argmin_{a \in \mathcal{A}}~ C_\mathrm{dist}(a \mid \hx) \quad  \text{subject to $H(\hx+a)=y^\star$},
\end{align*}
where
$\mathcal{A}$
is a set of feasible perturbation vectors, and $C_\mathrm{dist}$ is a cost function that measures the required efforts of $a$, such as TLPS~\cite{Ustun:FAT*2019} and DACE~\cite{Kanamori:IJCAI2020}.
\cref{tab:example:ce} presents examples of actions extracted from a logistic regression classifier trained on our credit approval dataset in~\cref{fig:demo}.
A user can obtain one's desired outcome by changing each feature according to the suggested perturbation vectors in \cref{tab:example:ce}.
 
In practice, however, showing only a perturbation vector $a^\star$ is often insufficient for users to execute the action due to interaction among features~\cite{Poyiadzi:AIES2020}.
In the previous example, as shown in the causal DAG in \cref{fig:demo}(b), we have an asymmetric interaction {``JobSkill'' $\stackrel{}{\to}$ ``Income"}, which means that increasing one's ``JobSkill" has a positive effect on increasing ``Income" while the opposite does not.
From these observations, we see that it is more reasonable to increase first ``JobSkill" and then ``Income" than the reverse order. 
Thus, practical CE methods are required to provide an appropriate order of changing features in addition to a perturbation vector $a^\star$. 

To achieve this requirement, we propose a novel CE framework that returns a pair $(a^\star,\sigma^\star)$, called an \emph{ordered action}, of a perturbation vector $a^\star$ and a permutation $\sigma^\star$ of features that advises a user to change features in that order.
We assume that the feature interaction is represented by an \emph{interaction matrix} $M$, whose element indicates the linear interaction between two features, such as correlations, causal effects, or given by some prior knowledge. 
Roughly speaking, we consider the following optimization problem: 
\begin{align*}
            (a^\star, \sigma^\star) :=  \argmin_{a \in \mathcal{A}, \sigma \in \Sigma}~ &C_\mathrm{dist}(a \mid \hx) + \gamma \cdot C_\mathrm{ord}(a, \sigma \mid M) \\
                                        \text{subject to}~ &H(\hx+a)=y^\star,
\end{align*}
where $C_\mathrm{ord}$ is a new cost function for determining an order of changing features, $\Sigma$ is the set of all permutations of the features perturbed by $a$, and $\gamma > 0$ is a trade-off parameter.

\subsection{Our Contributions}
Our goal is to extend CE framework so that it provides an ordered action by taking into account feature interaction.
Our contributions are summarized as follows: 
\begin{enumerate} 
    \item As a new framework for CE, we propose \emph{Ordered Counterfactual Explanation (OrdCE)} 
          that provides an \emph{ordered action}, i.e., a pair of a perturbation vector and an order of changing features.
          For that purpose, we introduce a new objective function that evaluates the cost of an ordered action based on a given \emph{interaction matrix} $M$. 
    \item We formulate the problem of finding an optimal ordered action as a \emph{mixed-integer linear optimization (MILO)} problem, which can be efficiently solved by modern MILO solvers such as CPLEX~\cite{cplex}. 
          Our formulation works on popular classifiers, such as linear models, tree ensembles, and multilayer perceptrons.
          In addition, the formulation can be combined with any existing cost function used in MILO-based CE methods, such as TLPS~\cite{Ustun:FAT*2019} or DACE~\cite{Kanamori:IJCAI2020}.
    \item We conducted numerical experiments on real datasets and compared the performance of our OrdCE with previous CE methods.
          We confirmed that 
          (i) our MILO approach obtained better ordered actions than baselines within practical computation time, and
          (ii) the obtained orders were reasonable from the perspective of feature interaction.
\end{enumerate}

\cref{tab:example:oce} presents examples of ordered actions extracted by OrdCE on the credit approval dataset in \cref{fig:demo}.
These orders of changing features are consistent with the causal DAG shown in \cref{fig:demo}(b). 
For example, the ordered action extracted by OrdCE + DACE indicates increasing ``WorkPerDay" before ``Income".
This order is more reasonable than its reverse order because ``WorkPerDay" has a positive effect on ``Income".
In addition, in the result of OrdCE + TLPS, the total number of the changing features increases from that of the unordered TLPS in \cref{tab:example:ce}. 
However, because ``JobSkill" has an effect $6.00$ on ``Income" as shown in \cref{fig:demo}(b), changing ``JobSkill" by $+1$ naturally causes an increase of ``Income" by $+6$.
Thus, it is expected that the user completes the ordered action suggested by OrdCE + TLPS with only changing ``JobSkill" at the 1st step. 
Our OrdCE can find such an appropriate order by optimizing a perturbation vector and an order simultaneously.
In summary, we see that our method presents a reasonable ordered action, which helps a user act to obtain the desired outcome.

\begin{table}[t]
    \centering
    \small
    \begin{tabular}{lccc}
        \toprule
        \multirow{2}{*}{\textbf{Method}} & \multicolumn{3}{c}{\textbf{Action}} \\ 
                                         & {``Income"} & {``WorkPerDay"} & {``HealthStatus"}  \\
        \toprule
        {DACE}                           & +4 & +1 & +3  \\
        \midrule
        {TLPS}                           & +5 &  0 & 0  \\
        \bottomrule
    \end{tabular}
    \caption{Examples of actions extracted by the existing CE methods on the credit approval dataset in~\cref{fig:demo}.}
    \label{tab:example:ce}
\end{table}
\begin{table}[t]
    \small
    \centering
    \begin{tabular}{lclc}
        \toprule
        \textbf{Method} & \textbf{Order} & \textbf{Feature} & \textbf{Action} \\ 
        \toprule
        \multirow{3}{*}{OrdCE + DACE}  & 1st & {``HealthStatus"}   & +3  \\ 
                                  & 2nd & {``WorkPerDay"}        & +1  \\ 
                                  & 3rd & {``Income"}         & +4   \\ 
        \midrule
        \multirow{2}{*}{OrdCE + TLPS}   & 1st & {``JobSkill"}        & +1  \\ 
                                  & 2nd & {``Income"}   & +6  \\ 
        \bottomrule
    \end{tabular}
    \caption{Examples of ordered actions extracted by our OrdCE on the credit approval dataset in \cref{fig:demo}.}
    \label{tab:example:oce}
\end{table}

\subsection{Related Work}
The existing CE methods can be categorized into gradient-based~\cite{Wachter:HJLT2018,Moore:PRICAI2019,Karimi:NIPS2020}, autoencoder~\cite{Dhurandhar:NIPS2018,Mahajan:NIPSWS2019}, SAT~\cite{Karimi:AISTATS2020}, or mixed-integer linear optimization (MILO)~\cite{Cui:KDD2015,Ustun:FAT*2019,Russell:FAT*2019,Kanamori:IJCAI2020}. 
Since our cost function is non-differentiable due to the discrete nature of a permutation $\sigma$ over features, we focus on MILO-based methods, which can directly handle such functions.



Most of CE methods provide only a perturbation vector as an action.
To the best of our knowledge, FACE~\cite{Poyiadzi:AIES2020} and OAS~\cite{Ramakrishnan:AAAI2020} are the only exceptions that consider a sequence of actions.
FACE provides a sequence of training instances as a path from a given instance to the target instances in a neighborhood graph. 
However, FACE does not take into account feature interaction and not determine the execution order of features.
On the other hands, OAS provides a sequence of actions, which is related to classical planning~\cite{Nau:2004:Planning}.  
There,  the costs of candidate actions are static and irrelevant to their order, while in OrdCE, the costs dynamically depend on the previously chosen actions due to their interaction. 
It is also noteworthy that \cite{Bertsimas:arxiv2019} studied how to determine an order of features to improve explainability in linear regression.

\section{Preliminaries} 

For a positive integer $n \in \mathbb{N}$, we write $[n] := \set{1,\dots,n}$.
For a proposition $\psi$, $\ind{\psi}$ denotes the indicator of $\psi$, i.e.,
$\ind{\psi}=1$ if $\psi$ is true, and $\ind{\psi}=0$ if $\psi$ is false.

Throughout this work, we consider a \emph{binary classification problem} as a prediction task, which is sufficient for CE. 
For a multi-class classification problem, we can reduce it to a binary one between the target class and the other classes.
We denote input and output domains by $\mathcal{X} = \mathcal{X}_1 \times \dots \times \mathcal{X}_D \subseteq \mathbb{R}^{D}$ and $\mathcal{Y} = \set{-1, +1}$, respectively.
An \emph{instance} is a vector $x = (x_1, \dots, x_D) \in \mathcal{X}$.
A \emph{classifier} is a function $H \colon \mathcal{X} \to \mathcal{Y}$.

\subsection{Additive Classifiers}
In this paper, we focus on CE with \emph{additive classifiers~(AC)} $H \colon \mathcal{X} \to \mathcal{Y}$ expressed in the following additive form~\cite{Hastie:2009:Elements}:
\begin{align*}
    H(x) = \mathrm{sgn}\left({\sum}_{t=1}^{T} w_t \cdot h_t(x) - b \right),
\end{align*}
where $T \in \mathbb{N}$ is the total number of base learners,
$h_t \colon \mathcal{X} \to \mathbb{R}$ is a base learner,
$w_t \in \mathbb{R}$ is a weight value of $h_t$ for $t \in [T]$,
and $b \in \mathbb{R}$ is an intercept.

\subsubsection{Linear Models}
\emph{Linear models (LM)}, such as Logistic Regression and Linear Support Vector Machines, are one of the most popular classifiers~\cite{Hastie:2009:Elements}.
We remark that an LM is a special case of AC such that $T = D$ and $h_d(x) = x_d$ for $d \in [D]$.

\subsubsection{Tree Ensembles}
\emph{Tree ensembles (TE)}, such as Random Forests~\cite{Breiman:ML2001},
are renowned for their high prediction performances in machine learning competitions.
Each base learner $h_t$ of a TE is a \emph{decision tree}, which is a classifier consisting of a set of if-then-else rules expressed in the form of a binary tree.
A decision tree $h_t \colon \mathcal{X} \to \mathcal{Y}$ with $L_t$ leaves represents a partition $r_{t,1}, \dots, r_{t,L_t}$ of the input domain $\mathcal{X}$~\cite{Hastie:2009:Elements}.
Then, it can be expressed as the linear combination $h_t(x) = \sum_{l=1}^{L_t} \hat{y}_{t,l} \cdot \ind{x \in r_{t,l}}$, where $\hat{y}_{t,l} \in \mathcal{Y}$ is a predictive label of the leaf $l \in [L_t]$.

\subsubsection{Multilayer Perceptrons}
\emph{Multilayer perceptrons~(MLP)}, or neural networks, have become increasingly more common over the past decade~\cite{Goodfellow:2016:Deep}.
For simplicity, we consider two-layer ReLU networks, i.e., MLP with one hidden layer and the rectified linear unit (ReLU) function $g(x) = \max(0,x)$ as an activation function.
Our proposed methods presented in~\Cref{sec:milo} can be extended to general multilayer ReLU networks.
In MLP, each base learner $h_t$ is an output of a neuron in its hidden layers with the ReLU function $g$.
It can be expressed as $h_t(x) = g(w^{(t)} \cdot x + b^{(t)})$, where $w^{(t)} \in \mathbb{R}^{D}$ and $b^{(t)} \in \mathbb{R}$ are weight values and an intercept with respect to the $t$-th neuron in the hidden layer, respectively.

\section{Problem Statement}
\subsection{Action and Ordered Action}
Let $H \colon \mathcal{X} \to \mathcal{Y}$ and $\hx = (\hx_1, \dots, \hx_D) \in \mathcal{X}$ be a classifier and a given instance such that $H(\hx)=-1$, respectively. 
We define an \emph{action} for $\hx$ as a perturbation vector $a \in \mathbb{R}^{D}$ such that $H(\hx+a)=+1$. 
An \emph{action set} $\mathcal{A} = A_1 \times \dots \times A_D$ is a finite set of feasible actions such that $0 \in A_d$ and $A_d \subseteq \{ a_d \in \mathbb{R} \mid \hx_d + a_d \in \mathcal{X}_d \}$ for $d \in [D]$. 
We can determine each $A_d$ depending on the type and constraint of the feature $d \in [D]$.
For example, a feature representing ``Age" must be a positive integer and cannot be decreased.
We define the \emph{perturbing features} of an action $a$ as $\supp{a} := \set{d \in [D] \mid a_d \not= 0}$.
For $K \in [D]$, we write $\mathcal{A}_{\leq K} := \set{a \in \mathcal{A} \mid |\supp{a}| \leq K}$.

We introduce an \emph{ordered action} for OrdCE.
An ordered action is a pair of a perturbation vector $a \in \mathcal{A}_{= K}$ for some $K \in [D]$ and a permutation $\sigma = (\sigma_1, \dots, \sigma_K) \in [D]^{K}$ of the perturbing features $\supp{a}$, which suggests perturbing the features $\supp{a}$ in that order.
We denote by $\Sigma(a)$ the set of all permutations of $\supp{a}$, and call $\sigma \in \Sigma(a)$ a \emph{perturbing order} of $a$.

\subsection{Interaction Matrix}\label{sec:interaction-matrix}
Practically, causal relationships are usually unknown in advance, and we need to estimate them. 
Since linear causal models can be estimated properly in practical settings where hidden common causes are included~\cite{Shimizu:JMLR2011}, we assume the feature interaction is linear.
We assume that the feature interaction is represented by a matrix $M=(M_{i,j})_{1 \leq i,j \leq D}$, which we call an \emph{interaction matrix}.
Each element $M_{i,j}$ represents the linear interaction from $i$ to $j$, that is, when we perturb a feature $x_i$ to $x_i + a_i$, then $x_j$ is perturbed to $x_j + M_{i,j} a_i$. 
We can compute $M_{i,j}$ explicitly with causal effect estimation methods~\cite{Pearl:2009,Shimizu:JMLR2011,Janzing:AS2013} or some prior knowledge of domain experts.

From a given causal DAG estimated by, for example, DirectLiNGAM~\cite{Shimizu:JMLR2011,Hyvarinen:JMLR2013}, we can compute an interaction matrix as follows.
Let $B = (B_{i,j})_{1 \leq i,j \leq D}$ be the adjacency matrix of the estimated causal DAG.
By reordering the order of the nodes, we can assume that $B$ is a strictly upper triangular matrix.
Here, LiNGAM considers a model expressed the following structural equations: 
$x_j = \sum_{i \in \operatorname{pa}_{B}(j)} B_{i,j} x_i + e_j$,
where $e_j$ is a continuous random variable that has non-Gaussian distributions and is independent of each other, and $\operatorname{pa}_{B}(j) \subseteq [D]$ is the set of features that are the ancestors of $j$ on the estimated causal DAG.
Then, we obtain
\begin{align*}
    M = I + {\sum}_{k = 1}^{D-1} B^k.
\end{align*}

\subsection{Cost Function}
As a score to evaluate the required effort of an ordered action $(a, \sigma)$, we introduce a new cost function $C_\mathrm{OrdCE}$ as follows. 
Given an input instance $\hx \in \mathcal{X}$,
an interaction matrix $M$, and a trade-off parameter $\gamma \geq 0$, 
we define $C_\mathrm{OrdCE}$ for a pair of a perturbation $a \in \mathcal{A}$ and its order $\sigma \in \Sigma(a)$ as
\begin{align*}
    C_\mathrm{OrdCE}&(a, \sigma \mid \hx, M, \gamma) \\
    &:= C_\mathrm{dist}(a \mid \hx) + \gamma \cdot C_\mathrm{ord}(a, \sigma \mid M),    
\end{align*}
where $C_\mathrm{dist}$ and $C_\mathrm{ord}$ are \emph{distance-based} and \emph{ordering cost functions}, respectively.
The former evaluates the required effort of a perturbation vector $a$, and the latter determines a perturbing order $\sigma$ of $a$. 

\subsubsection{Distance-based Cost Function}
As with the existing CE methods, we utilize a distance-based cost function $C_\mathrm{dist}$ to evaluate the required effort of an entire perturbation $a$.
For simplicity, we assume $C_\mathrm{dist}$ as the following form: 
\begin{align*}
    C_\mathrm{dist}(a \mid \hx) = {\sum}_{d=1}^{D} \mathrm{dist}_{d}(a_{d} \mid \hx_{d}),
\end{align*}
where $\mathrm{dist}_d \colon A_d \to \bbR_{\geq 0}$ is a cost measure of the feature $d$ 
that evaluates the effort to change $\hx_d$ to $\hx_d+a_d$, such as the total-log percentile shift~\cite{Ustun:FAT*2019}.
Note that our optimization approach can adapt to other types of existing cost functions,
such as $\ell_1$-Mahalanobis' distance~\cite{Kanamori:IJCAI2020}.
While these useful distance-based cost functions have been proposed, they do not deal with a perturbing order $\sigma \in \Sigma(a)$.

\subsubsection{Ordering Cost Function}
To extend CE so as to deal with a perturbing order $\sigma = (\sigma_1, \dots, \sigma_K) \in \Sigma(a)$, we introduce an ordering cost function $C_\mathrm{ord}$ as the following form: 
\begin{align*}
    C_\mathrm{ord}(a, \sigma \mid M) = {\sum}_{k=1}^{K} \mathrm{cost}^{(k)}(a_{\sigma_1,\dots,\sigma_k} \mid M), 
\end{align*}
where $a_{\sigma_1, \dots, \sigma_k} := (a_{\sigma_1}, \dots, a_{\sigma_k})$ and $\mathrm{cost}^{(k)}$ is a cost for each $k$-th perturbation $a_{\sigma_k}$.
We define $\mathrm{cost}^{(k)}$ as a function that depends not only on the perturbation of $\sigma_k$ at the $k$-th step but also on the previously perturbed features ${\sigma_1}, \dots, {\sigma_{k-1}} \in \supp{a}$ since the actual amount of a perturbation of a feature is affected by the values of other features that interact with it. 
Note that we assume that the previously perturbed features are unaffected by the following perturbation, which is based on the \emph{intervention} in causal  models~\cite{Pearl:2009}.

To define $\mathrm{cost}^{(k)}$, we introduce
a parameter
$\Delta^{(k)}=\Delta^{(k)}(a_{\sigma_1, \dots, \sigma_k} \mid M)$, 
called the \emph{actual perturbation}, as the amount of change on $x_{\sigma_k}$ that we actually need to obtain a desired perturbation. 
Then, the resulting perturbation $a_{\sigma_k}$ is equal to the sum of $\Delta^{(k)}$ and the effect of the previous perturbations $\Delta^{(1)}, \dots, \Delta^{(k-1)}$. 
Formalizing this idea, we have the following
conditions
on $\Delta^{(k)}$
for $k=1,2,3$: 
\begin{align*}
    a_{\sigma_1} & = \Delta^{(1)}, \\
    a_{\sigma_2} & = \Delta^{(2)} + M_{\sigma_1, \sigma_2} \cdot \Delta^{(1)}, \\
    a_{\sigma_3} & = \Delta^{(3)} + M_{\sigma_2, \sigma_3} \cdot \Delta^{(2)} + M_{\sigma_1, \sigma_3} \cdot \Delta^{(1)}.
\end{align*}
Generally, we have 
\begin{math}
a_{\sigma_k} = \Delta^{(k)} + {\sum}_{l=1}^{k-1} M_{\sigma_l, \sigma_k} \cdot \Delta^{(l)}.  
\end{math}
For any $k \in [K]$, we inductively obtain
\begin{align*}
    \Delta^{(k)}&(a_{\sigma_1, \dots, \sigma_k} \mid M) \\
    &= a_{\sigma_k} - {\sum}_{l=1}^{k-1} M_{\sigma_l, \sigma_k} \cdot \Delta^{(l)}(a_{\sigma_1, \dots, \sigma_l} \mid M).
\end{align*}
By using $\Delta^{(k)}$, we define $\mathrm{cost}^{(k)}$ as follows:
\begin{align*}
    \mathrm{cost}^{(k)}(a_{\sigma_1,\dots,\sigma_k} \mid M) := \left| \Delta^{(k)}(a_{\sigma_1, \dots, \sigma_k} \mid M) \right|.
\end{align*}
In practice, since each feature has different scale, we multiply each $\Delta^{(k)}$ by a scaling factor $s_{\sigma_{k}} > 0$, such as the inverse of its standard deviation.

\subsection{Problem Definition}
Our aim is to find an ordered action $(a,\sigma)$ that minimizes the cost $C_\mathrm{OrdCE}$.
This problem can be defined as follows.

\begin{problem}\label{prob:ordce}
Given an additive classifier $H \colon \mathcal{X} \to \mathcal{Y}$, an input instance $\hx \in \mathcal{X}$ such that $H(\hx)=-1$, an action set $\mathcal{A}$, an interaction matrix $M \in \mathbb{R}^{D \times D}$, $K \in [D]$, and $\gamma \geq 0$, 
find an ordered action $(a,\sigma)$ that is an optimal solution for the following optimization problem:
\begin{align*}
    \begin{array}{cc}
        \displaystyle \minimize_{a \in \mathcal{A}_{\leq K}, \sigma \in \Sigma(a)} & C_\mathrm{OrdCE}(a,\sigma \mid \hx, M, \gamma)  \\
        \text{subject to} & H(\hx+a) = +1. 
    \end{array}
\end{align*}
\end{problem}

By solving the above optimization problem, we obtain an ordered action $(a,\sigma)$ that accords with feature interaction.

\subsection{Concrete Examples}
\label{appendix:def_conc_example}
To study the cost function $C_\mathrm{ord}$ and objective function $C_\mathrm{OrdCE}$, we present concrete examples to observe behavior of these functions in the same setting as \Cref{sec:intro}. 
Our synthetic credit loan approval dataset consists of five features $x_1, \dots x_5$ presented in \cref{fig:demo} in \Cref{sec:intro}. 
As an interaction matrix, we take the matrix $M$ whose element $M_{i,j}$ represents the average causal effect from a feature $i$ to $j$. 
As described previously, $M = (M_{i,j})_{1 \leq i,j \leq 5}$ is calculated from the adjacency matrix of the causal DAG in \cref{fig:demo}(b) as follows:
\begin{align*}
    M = 
    \bordermatrix{
        & 1 & 2 & 3 & 4 & 5  \cr 
        1 & 1 & 1 & 6 & 0 & 0  \cr
        2 & 0 & 1 & 6 & 0 & 0  \cr
        3 & 0 & 0 & 1 & 0 & 0  \cr
        4 & 0 & 0 & 4 & 1 & -0.5  \cr
        5 & 0 & 0 & 0 & 0 & 1  \cr
    }.
\end{align*}
In the examples below, we use $C_\mathrm{ord}$ with scaling factors $s_d > 0 \ (d \in [D])$: 
\begin{align*}
    C_\mathrm{ord}(a, \sigma \mid M) = {\sum}_{k=1}^{K} s_{\sigma_k} \cdot \left| \Delta^{(k)}(a_{\sigma_1, \dots, \sigma_k} \mid M) \right|.
\end{align*}

First, we show an example to observe behavior of the ordering cost function $C_\mathrm{ord}$ in the following \Cref{example:ord}.

\begin{example}\label{example:ord}
Consider a  perturbation $a = (0, 0, 4, 1, 3)$ and its  perturbing orders $\sigma = (4,3,5)$ and $\sigma^\circ = (5,4,3)$.
We compare the values of our ordering cost function $C_\mathrm{ord}$ for the two ordered actions $(a,\sigma)$ and $(a, \sigma^\circ)$.

For $(a,\sigma)$, the actual perturbations $\Delta^{(k)}$ can be calculated as follows:
\begin{align*}
    &\Delta^{(1)} = 1 - 0 = 1, \\
    &\Delta^{(2)} = 4 - M_{4,3} \cdot \Delta^{(1)} = 0, \\
    &\Delta^{(3)} = 3 - M_{3,5} \cdot \Delta^{(2)} - M_{4,5} \cdot \Delta^{(1)} = 3.5.
\end{align*}
Thus, the value of $C_\mathrm{ord}$ for $(a,\sigma)$ can be calculated as
\begin{align*}
    C_\mathrm{ord}(a, \sigma \mid M)  
    &= s_{\sigma_1} \cdot |\Delta^{(1)}| + s_{\sigma_2} \cdot |\Delta^{(2)}| + s_{\sigma_3} \cdot |\Delta^{(3)}| \\
    &= s_4 + 3.5 s_5.
\end{align*}

Similarly, 
the value of $C_\mathrm{ord}$ for $(a,\sigma^\circ)$ can be calculated as 
\begin{align*}
    C_\mathrm{ord}(a, \sigma^\circ \mid M)  
    &= s_{\sigma^\circ_1} \cdot |\Delta^{(1)}| + s_{\sigma^\circ_2} \cdot |\Delta^{(2)}| + s_{\sigma^\circ_3} \cdot |\Delta^{(3)}| \\
    &= s_4 + 3 s_5.
\end{align*} 

Because $s_d > 0$ for all $d \in \set{1,\dots,5}$, $C_\mathrm{ord}(a, \sigma \mid M) > C_\mathrm{ord}(a, \sigma^\circ \mid M)$ holds.
\myendbox
\end{example}

In the above example, the ordered action $(a,\sigma)$ suggests increasing the values of ``WorkPerDay", ``Income", and ``HealthStatus" in this order.
The other ordered action $(a,\sigma^\circ)$ suggests increasing the values of ``HealthStatus", ``WorkPerDay", and ``Income" in this order.
Since ``WorkPerDay" has a negative causal effect to ``HealthStatus", the perturbing order $\sigma^\circ$ is more reasonable than $\sigma$ from the perspective of the feature interaction.
The above example indicates that we can obtain an appropriate order of its perturbing features by minimizing $C_\mathrm{ord}$.

Next, we show an example to observe behavior of the objective cost function $C_\mathrm{OrdCE}$ in the following \Cref{example:obj}.

\begin{example}\label{example:obj}
Consider the same setting as Example~\ref{example:ord} and two feasible ordered actions $(a,\sigma)$ and $(a^\circ, \sigma^\circ)$ with 
\begin{align*}
    \begin{array}{ll}
        a = (0,0,6,0,0), &\sigma = (3), \\
        a^\circ = (0,1,6,0,0), &\sigma^\circ = (2,3).
    \end{array} 
\end{align*}
We compare the values of our objective function $C_\mathrm{OrdCE}$ for the two ordered actions $(a,\sigma)$ and $(a^\circ, \sigma^\circ)$.

For $(a,\sigma)$, the actual perturbations $\Delta^{(k)}$ can be calculated as follows:
\begin{align*}
    \Delta^{(1)} = 6 - 0 = 6.
\end{align*}
Thus, the value of $C_\mathrm{OrdCE}$ for $(a,\sigma)$ can be calculated as 
\begin{align*}
    C_\mathrm{OrdCE}(a, \sigma \mid \hx, M, \gamma) 
    &= c_3 + \gamma \cdot ( s_{\sigma_1} \cdot |\Delta^{(1)}| ) \\
    &= c_3 + \gamma \cdot 6 s_3,
\end{align*}
where $c_3 = \mathrm{dist}_{3}(6 \mid \hx_3) > 0$. 

Similarly, 
the value of $C_\mathrm{OrdCE}$ for $(a^\circ,\sigma^\circ)$ can be calculated as 
\begin{align*}
    &C_\mathrm{OrdCE}(a^\circ, \sigma^\circ \mid \hx, M, \gamma) \\
    &= c_2 + c_3 + \gamma \cdot ( s_{\sigma^\circ_1} \cdot |\Delta^{(1)}| + s_{\sigma^\circ_2} \cdot |\Delta^{(2)}|) \\
    &= c_2 + c_3 + \gamma \cdot s_2,
\end{align*}
where $c_2 = \mathrm{dist}_{2}(1 \mid \hx_2) > 0$.

Now we assume $6s_3 - s_2 > 0$.
Then, we obtain the following result:
\begin{align*}
    &C_\mathrm{OrdCE}(a, \sigma \mid \hx, M, \gamma) > C_\mathrm{OrdCE}(a^\circ, \sigma^\circ \mid \hx, M, \gamma) \\
    &\iff \gamma \cdot (6s_3 - s_2) - c_2 > 0 \\
    &\iff \gamma > \frac{c_2}{6s_3 - s_2}.
\end{align*}
Hence, if $6s_3 - s_2 > 0$ and $\gamma > {c_2}/{(6s_3 - s_2)}$, 
then $C_\mathrm{OrdCE}(a, \sigma \mid \hx, M, \gamma) > C_\mathrm{OrdCE}(a^\circ, \sigma^\circ \mid \hx, M, \gamma)$. 
\myendbox
\end{example}

In the above example, the ordered action $(a,\sigma)$ suggests increasing only ``Income".
The other ordered action $(a^\circ, \sigma^\circ)$ suggests increasing the values of ``JobSkill" and ``Income" in this order.
The total number of the changing features of the latter ordered action is greater than that of the former.
However, a user is expected to complete the latter ordered action with only changing ``JobSkill" since increasing one’s ``JobSkill” has a positive effect on increasing ``Income” as mentioned in \Cref{sec:intro}.
From the above example, we see that we can adjust a trade-off between the required effort of an entire perturbation $a$ and each step $\Delta^{(k)}$ by tuning the parameter~$\gamma$. 

\section{Optimization Framework}\label{sec:milo}
In this section, we formulate our problem for finding an optimal ordered action as an MILO problem. 

\subsection{Basic Constraints}
For $d \in [D]$, we set $A_d = \set{a_{d,1}, \dots, a_{d,I_d}}$ and $a_{d,1}=0$.
First, we introduce binary variables $\pi_{d,i}\in \{0,1\}$ for $d \in [D]$ and $i\in [I_d]$, 
which indicate that $a_{d,i}\in A_d$ is selected $(\pi_{d,i}=1)$ or not $(\pi_{d,i}=0)$ 
as in the previous MILO-based methods~\cite{Ustun:FAT*2019,Kanamori:IJCAI2020}.
Then, $\pi_{d,i}$ must satisfy the following constraints:
\begin{align}\label{const:oneaction}
    & {\sum}_{i=1}^{I_d} \pi_{d,i} = 1, \forall d \in [D].
\end{align}
By using $\pi_{d,i}$, we can express a perturbation for each feature $d$ as $a_d = \sum_{i=1}^{I_d} a_{d,i} \cdot \pi_{d,i}$.
Note that $\pi_{d,1}=1$ indicates that a feature $d$ is not perturbed since $a_{d,1}=0$.

To express an order of perturbing features, 
we introduce binary variables $\pi^{(k)}_{d,i} \in \{0,1\}$ for $d \in [D], i \in [I_d]$, and $k\in [K]$, 
which indicate $a_{d,i} \in A_d$ is selected as the $k$-th perturbation; 
that is, $\pi^{(k)}_{d,i}=1$ if $a_{d,i} \in A_d$ is selected in the $k$-th step, and $\pi^{(k)}_{d,i}=0$ otherwise.
Then, $\pi^{(k)}_{d,i}$ must satisfy the following constraints:
\begin{align}\label{const:oneactionforeachk}
    &\pi_{d,i} = {\sum}_{k=1}^{K} \pi^{(k)}_{d,i}, \forall d \in [D], \forall i \in [I_d].
\end{align}
We also introduce binary variables $\sigma_{k,d}$ for $k \in [K]$ and $d \in [D]$ that indicate whether the feature $d$ is perturbed in the $k$-th step.
We impose the following constraints on $\sigma_{k,d}$: 
\begin{align}
    &\sigma_{k,d} = 1 - \pi^{(k)}_{d,1}, \forall k \in [K], \forall d \in [D], \label{const:sigma1} \\
    &{\sum}_{k=1}^{K} \sigma_{k,d} \leq 1, \forall d \in [D], \label{const:sigma2} \\
    &{\sum}_{d=1}^{D} \sigma_{k,d} \leq 1, \forall k \in [K], \label{const:sigma3} \\
    &{\sum}_{d=1}^{D} \sigma_{k,d} \geq {\sum}_{d=1}^{D} \sigma_{k+1,d}, \forall k \in [K-1]. \label{const:sigma4}
\end{align}
\pref{const:sigma2} allows that we can perturb each feature at most once.
\pref{const:sigma3} imposes that we can perturb at most one feature in each step.
\pref{const:sigma4} is a symmetry breaking constraint on $\sigma_{k,d}$.

\subsection{Objective Function}
Our objective function $C_\mathrm{OrdCE}$ consists of $C_\mathrm{dist}$ and $C_\mathrm{ord}$, which we express with the program variables $\pi^{(k)}_{d,i}$ and $\sigma_{k,d}$.

\subsubsection{Distance-based Cost Function}
From our assumption of $C_\mathrm{dist}$, it can be expressed as follows:
\begin{align*}
    C_\mathrm{dist}(a \mid \hx) = {\sum}_{d=1}^{D} {\sum}_{i=1}^{I_d} {\sum}_{k=1}^{K} c_{d,i} \cdot \pi^{(k)}_{d,i},
\end{align*}
where $c_{d,i}$ is a constant value such that $c_{d,i} = \mathrm{dist}_d(a_{d,i} \mid \hx_d)$.
Note that our MILO formulation can adapt to other existing cost functions
such as DACE~\cite{Kanamori:IJCAI2020} and SCM~\cite{Mahajan:NIPSWS2019}.

\subsubsection{Ordering Cost Function}
Since $C_\mathrm{ord}$ is non-linear due to a permutation $\sigma$, we need to express it by linear constraints of the variables.
We introduce variables $\zeta_k$ for $k \in [K]$ such that $\zeta_k = | \Delta^{(k)} |$.
Then, $C_\mathrm{ord}$ can be expressed as follows:
\begin{align*}
    C_\mathrm{ord}(a, \sigma \mid M) = {\sum}_{k=1}^{K} \zeta_{k}.
\end{align*}
Moreover, for $k \in [K]$ and $d \in [D]$, we introduce variables $\delta_{k,d} \in \bbR$ to express $\zeta_k = |\sum_{d=1}^{D} \delta_{k,d}|$.
Then, $\delta_{k,d}$ must satisfy $\delta_{k,d}=\Delta^{(k)}$ if $\sigma_{k,d}=1$, and $\delta_{k,d}=0$ if $\sigma_{k,d}=0$.
We can linearize these non-linear constraints as follows:
\begin{align}
    &\delta_{k,d} \geq {\sum}_{i=1}^{I_d} a_{d,i} \cdot \pi^{(k)}_{d,i} - \epsilon_{k,d} - U_{k,d} \cdot (1-\sigma_{k,d}), \notag \\
    &\hspace{46.5mm}\forall k \in [K], \forall d \in [D], \label{const:ordering1} \\
    &\delta_{k,d} \leq {\sum}_{i=1}^{I_d} a_{d,i} \cdot \pi^{(k)}_{d,i} - \epsilon_{k,d} - L_{k,d} \cdot (1-\sigma_{k,d}), \notag \\
    &\hspace{46.5mm}\forall k \in [K], \forall d \in [D], \label{const:ordering2}  \\
    &L_{k,d} \cdot \sigma_{k,d} \leq \delta_{k,d} \leq U_{k,d} \cdot \sigma_{k,d}, \forall k \in [K], \forall d \in [D], \label{const:ordering3}  \\
    &\epsilon_{k,d} = {\sum}_{l=1}^{k-1}{\sum}_{d'=1}^{D} M_{d',d} \cdot \delta_{l,d'}, \notag \\
    &\hspace{45mm}\forall k \in [K], \forall d \in [D], \label{const:ordering4}  \\
    &-\zeta_{k} \leq {\sum}_{d=1}^{D} \delta_{k,d} \leq \zeta_{k},  \forall k \in [K], \label{const:ordering5} 
\end{align}
where $\epsilon_{k,d}$ is an auxiliary variable such that $\epsilon_{k,d} = \sum_{l=1}^{k-1} M_{\sigma_l, d} \cdot \Delta^{(l)}$ for $k \in [K]$ and $d \in [D]$.
The constant values $L_{k,d}$ and $U_{k,d}$ are the lower and upper bounds of $\delta_{k,d}$.
These values can be recursively computed from the interaction matrix $M$ and the action set $\mathcal{A}$ as follows:
\begin{align*}
    L_{k+1,d} = L_{k,d} - \max_{d' \in [D] \setminus \set{d}} \max_{\Delta \in \set{L_{k,d'}, U_{k,d'}}} M_{d',d} \cdot \Delta, \\
    U_{k+1,d} = U_{k,d} - \min_{d' \in [D] \setminus \set{d}} \min_{\Delta \in \set{L_{k,d'}, U_{k,d'}}} M_{d',d} \cdot \Delta, 
\end{align*}
where $L_{1,d}=\min_{a_d \in A_d} a_{d}$ and $U_{1,d}=\max_{a_d \in A_d} a_{d}$.

\subsection{Base Learner Constraints}
We introduce variables $\xi_t \in \mathbb{R}$ for $t \in [T]$ such that $\xi_t = h_t(\hx+a)$, where $h_t$ is the $t$-th base learner of $H$.
From the definition of additive classifiers, the constraint $H(\hx+a) = +1$ is equivalent to the following linear constraint of $\xi_t$:
\begin{align}\label{const:decisionfunction}
    {\sum}_{t=1}^{T} w_t \cdot \xi_t \geq b.
\end{align}
We express the constraint $\xi_t = h_t(\hx+a)$ by linear constraints of $\xi_t$ and $\pi_{d,i}$ because $h_t(\hx+a)$ depends on the value of $a$, i.e., the variables $\pi_{d,i}$. 
In the following, we show how to express $\xi_t$ when $H$ is a linear model~(LM), tree ensemble~(TE), or multilayer perceptron~(MLP).

\subsubsection{Linear Models}
From the definition of LM, $T=D$ and $h_d(\hx + a) = \hx_d + a_d$ for $d \in [D]$.
Hence, we can simply express the base learner of the LM as follows:
\begin{align}\label{const:linear}
    \xi_d = \hx_d + {\sum}_{i=1}^{I_d} a_{d,i} \cdot \pi_{d,i}, \forall d \in [D].
\end{align}

\subsubsection{Tree Ensembles}
Each base learner $h_t$ of the TE is a decision tree.
To express $\xi_t = h_t(\hx+a)$, we can utilize the following \emph{decision logic constraint}~\cite{Cui:KDD2015,Kanamori:IJCAI2020}:
\begin{align}
    &\phi_{t,l} \in \set{0,1}, \forall t \in [T], \forall l \in [L_t], \label{const:leafind} \\
    &{\sum}_{l=1}^{L_t} \phi_{t,l} = 1, \forall t \in [T],  \label{const:leaf} \\
    &D \cdot \phi_{t,l} \leq {\sum}_{d=1}^{D} {\sum}_{i \in I_{t,l}^{(d)}} \pi_{d,i}, \forall t \in [T], \forall l \in [L_t], \label{const:decisionlogic} \\
    &\xi_t = {\sum}_{l=1}^{L_t} {\hat y}_{t,l} \cdot \phi_{t,l}, \forall t \in [T], \label{const:tree}
\end{align}
where $I_{t,l}^{(d)} = \set{i \in [I_{d}] \mid \hx_{d} + a_{d,i} \in r_{t,l}^{(d)}}$ and
$r_{t,l}^{(d)}$ is the subspace of $\mathcal{X}_d$ such that $r_{t,l} = r_{t,l}^{(1)} \times \dots \times r_{t,l}^{(D)}$.

\subsubsection{Multilayer Perceptrons}
Each base learner $h_t$ of the MLP is an output of the $t$-th neuron with the ReLU activation function,
i.e., $h_t(x+a) = \max \set{0,w^{(t)}(x+a)+b^{(t)}}$.
Hence, we need to extract the positive part of $w^{(t)}(x+a)+b^{(t)}$ as the output of the $t$-th base learner $\xi_t$.
To express it, we can utilize the following constraints proposed by~\cite{Serra:ICML2018}:
\begin{align}
    &\nu_t \in \set{0,1}, \bar{\xi}_t \geq 0, \forall t \in [T], \label{const:mlpvariables} \\
    &{\xi}_t \leq H_t \cdot \nu_t, \forall t \in [T],  \label{const:neuronpositive} \\
    &\bar{\xi}_t \leq \bar{H}_t \cdot (1-\nu_t), \forall t \in [T],  \label{const:neuronnegative} \\
    &\xi_t = \bar{\xi}_t + {\sum}_{d=1}^{D} w^{(t)}_d {\sum}_{i=1}^{I_d} a_{d,i} \cdot \pi_{d,i} + F_{t}, \forall t \in [T], \label{const:mlp}
\end{align}
where $F_t$, $H_t$, and $\bar{H}_t$ are constants such that $F_t = w^{(t)} \hx + b^{(t)}$, $H_t \geq \max_{a \in \mathcal{A}} w^{(t)} (x+a)+b^{(t)}$, and $\bar{H}_t \geq - \min_{a \in \mathcal{A}} w^{(t)} (x+a)+b^{(t)}$.
The variable $\nu_t$ indicates whether $w^{(t)}(x+a)+b^{(t)}$ is positive,
and $\bar{\xi}_t$ represents negative part of $w^{(t)}(x+a)+b^{(t)}$.
Note that the above constraints can be extended to a general MLP with more than two hidden layers~\cite{Serra:ICML2018}.

\subsection{Overall Formulation}
Finally, we show our overall formulation as follows:
\renewcommand{\arraystretch}{1.5}
\begin{align}\label{ordce:milo}
    \begin{array}{cl}
        \text{minimize} & \sum_{d=1}^{D}\sum_{i=1}^{I_d}\sum_{k=1}^{K} c_{d,i} \pi^{(k)}_{d,i} + \gamma \cdot \sum_{k=1}^{K} \zeta_k \\
        \text{subject to} & \text{Constraint~(\ref{const:oneaction}--\ref{const:decisionfunction})}, \\
                          & \begin{cases}
                                \text{\pref{const:linear}}, &\text{if } H \text{ is a LM,} \\
                                \text{Constraint~(\ref{const:leafind}--\ref{const:tree})},  &\text{if } H \text{ is a TE,} \\
                                \text{Constraint~(\ref{const:mlpvariables}--\ref{const:mlp})}, &\text{if } H \text{ is a MLP,}
                            \end{cases} \\
                          & \pi^{(k)}_{d,i} \in \set{0,1}, \forall k \in [K], \forall d \in [D], \forall i \in [I_d], \\
                          & \sigma_{k,d} \in \set{0,1}, \forall k \in [K], \forall d \in [D], \\
                          & \delta_{k,d}, \zeta_{k} \in \mathbb{R}, \forall k \in [K], \forall d \in [D]. 
    \end{array}
\end{align}
\renewcommand{\arraystretch}{1.0}
As with the existing MILO-based methods~\cite{Ustun:FAT*2019,Russell:FAT*2019,Kanamori:IJCAI2020}, 
our formulation can be
(i)~handled by off-the-shelf MILO solvers, such as CPLEX \cite{cplex}, and
(ii)~customized by additional user-defined constraints, such as one-hot encoded categorical features and hard constraints for ordering (e.g., \emph{precondition}~\cite{Ramakrishnan:AAAI2020}).
In summary, we can obtain ordered actions that satisfy user-defined constraints without implementing designated algorithms.

For computational complexity, Problem~\eqref{ordce:milo} includes $K$ times more variables and constraints than in the unordered one. 
Thus, solving \eqref{ordce:milo} is equal to or mode difficult than unordered ones.
However, in the context of CE, sparse actions are preferred from the perspective of interpretability~\cite{Wachter:HJLT2018}.
Therefore, it is sufficient to choose small $K$ for obtaining sparse actions.

\subsection{Post-processing and Partially Ordered Actions}

When some perturbing features have no interaction, changing the order of such features does not affect the cost $C_{\mathrm{ord}}$. 
For example in \cref{fig:demo}(b), the cost of the ordered action [``Education" $\to$ ``WorkPerDay" $\to$ ``JobSkill"] is the same as that of [``WorkPerDay" $\to$ ``Education" $\to$  ``JobSkill"] because ``WorkPerDay" has no effect to the others. 
Thus the ordered action can be reduced to the partially ordered action [``WorkPerDay"] and [``Education" $\to$ ``JobSkill"].
Suggesting such a partial order helps a user to execute an ordered action. 
We provide a post-processing algorithm that computes a partial order of the perturbing features from an ordered action and an interaction matrix.

An ordered action may be reduced to a \emph{partially ordered action}, which is a pair $(a, \le)$ of a perturbation vector $a \in \mathcal{A}$ and a partial order $\le$ on $\supp{a}$.
Here, we give a procedure to construct a partially ordered structure $\le$ from an interaction matrix $M$ and an ordered action $(a,\sigma)$. 
If a perturbing order $\sigma'$ of $a$ is consistent with the obtained partial order $\le$, the ordered action $(a,\sigma')$ has the same ordering cost $C_\mathrm{ord}$ with that of $(a,\sigma)$.

\begin{figure}[t]
    \centering
    \begin{tikzpicture}
        
        \node[draw, rectangle,rounded corners=3pt] (a) at (0,0) {\small 3};
        \node[draw, rectangle,rounded corners=3pt] (b) at (1,0) {\small 4};
        \node[draw, rectangle,rounded corners=3pt] (c) at (2,0) {\small 1};
        \node[draw, rectangle,rounded corners=3pt] (d) at (3,0) {\small 2};
        \node[draw, rectangle,rounded corners=3pt] (e) at (4,0) {\small 6};
        
        \draw[->] (a)--(b) node [midway, above left] {\small };
        \draw[->] (b)--(c) node [midway, above right] {\small };
        \draw[->] (c)--(d) node [midway, below left] {\small };
        \draw[->] (d)--(e) node [midway, below left] {\small };
        \node at (2,-0.8) {(a): Path structure for a given ordered action};

        \node[draw, rectangle,rounded corners=3pt] (a) at (0,-2) {\small 3};
        \node[draw, rectangle,rounded corners=3pt] (b) at (1,-2) {\small 4};
        \node[draw, rectangle,rounded corners=3pt] (c) at (2,-2) {\small 1};
        \node[draw, rectangle,rounded corners=3pt] (d) at (3,-2) {\small 2};
        \node[draw, rectangle,rounded corners=3pt] (e) at (4,-2) {\small 6};
        \node at (2,-3) {(b): Transitive closure of (a)};
        \draw[->] (a)--(b) node [midway, above left] {\small };
        \draw[->] (b)--(c) node [midway, above right] {\small };
        \draw[->] (c)--(d) node [midway, below left] {\small };
        \draw[->] (d)--(e) node [midway, below left] {\small };

        \draw[bend right,distance=0.7cm, ->] (a) to (c) node [midway, below left] {\small };
        \draw[bend right,distance=1.2cm, ->] (a) to (d) node [midway, below left] {\small };
        \draw[bend right,distance=1.7cm, ->] (a) to (e) node [midway, below left] {\small };
        \draw[bend left,distance=0.7cm, ->] (b) to (d) node [midway, below left] {\small };
        \draw[bend left,distance=1.2cm, ->] (b) to (e) node [midway, below left] {\small };
        \draw[bend left,distance=0.7cm, ->] (c) to (e) node [midway, below left] {\small };
        \node[draw, rectangle,rounded corners=3pt] (a) at (0,-4) {\small 3};
        \node[draw, rectangle,rounded corners=3pt] (b) at (1,-4) {\small 4};
        \node[draw, rectangle,rounded corners=3pt] (c) at (2,-4) {\small 1};
        \node[draw, rectangle,rounded corners=3pt] (d) at (3,-4) {\small 2};
        \node[draw, rectangle,rounded corners=3pt] (e) at (4,-4) {\small 6};
        \node at (2,-5) {(c): Remove the edges with no interaction  from (b)};
        \draw[->] (b)--(c) node [midway, above right] {\small };
        \draw[->] (c)--(d) node [midway, below left] {\small };
        \draw[bend right,distance=0.7cm, ->] (a) to (c) node [midway, below left] {\small };
        \draw[bend right,distance=1.2cm, ->] (a) to (d) node [midway, below left] {\small };
        \draw[bend left,distance=1.2cm, ->] (b) to (e) node [midway, below left] {\small };
        \draw[bend left,distance=0.7cm, ->] (c) to (e) node [midway, below left] {\small };

        \node[draw, rectangle,rounded corners=3pt] (a) at (0,-6) {\small 3};
        \node[draw, rectangle,rounded corners=3pt] (b) at (1,-6) {\small 4};
        \node[draw, rectangle,rounded corners=3pt] (c) at (2,-6) {\small 1};
        \node[draw, rectangle,rounded corners=3pt] (d) at (3,-6) {\small 2};
        \node[draw, rectangle,rounded corners=3pt] (e) at (4,-6) {\small 6};
        \node at (2,-6.8) {(d): Transitive reduction of (c)};
        \draw[->] (b)--(c) node [midway, above right] {\small };
        \draw[->] (c)--(d) node [midway, below left] {\small };
        \draw[bend right,distance=0.7cm, ->] (a) to (c) node [midway, below left] {\small };
         \draw[bend left,distance=0.7cm, ->] (c) to (e) node [midway, below left] {\small };

    \end{tikzpicture}
    \caption{Algorithm for obtaining a partially ordered action.
             }
    \label{fig:algorithm:partial}
\end{figure}
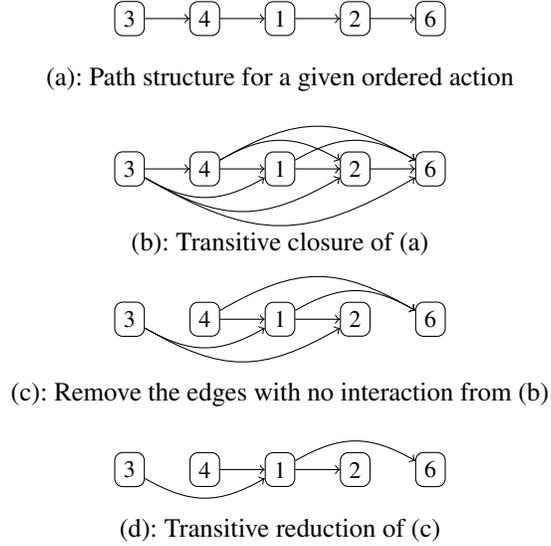

An ordered action can be expressed in the form of a path structure like \cref{fig:algorithm:partial}(a), where each node indicates a perturbing feature. 
In this path structure, changing a feature~$i$ should be executed after changing all its ancestors $\mathrm{pa}(i)$ and not after any its descendant. 
A partially ordered action is expressed in the form of a DAG like \cref{fig:algorithm:partial}(d), and a change in a feature satisfies the same condition as above. 
Note that even if a causal DAG is given, the DAG of a partially ordered action is not necessarily a subgraph of the causal DAG.

\subsubsection{Algorithm}

We give an algorithm to obtain a partially ordered action from an ordered action and an interaction matrix $M$. 
The procedure is as follows:
\begin{enumerate}
    \item Construct a path that represents the perturbing order of a given ordered action.
    \item Compute the transitive closure that represents the total order of the perturbing features. 
    \item Remove an edge from $i$ to $j$ if there is no interaction between $i$ and $j$, that is, $M_{i, j} = M_{j, i} = 0$. 
    \item Compute the transitive reduction that represents a partially ordered action.
\end{enumerate}
Here, a \emph{transitive closure} of a directed graph $G$ is a directed graph that has an edge from $i$ to $j$ if and only if there is a directed path from $i$ to $j$ in $G$.
Also, a \emph{transitive reduction} of a directed graph $G$ is a directed graph with the fewest number of edges whose transitive closure is the same as the transitive closure of $G$. 
For a finite DAG, its transitive closure and its transitive reduction are uniquely determined and can be computed in polynomial time~\cite{Munro:IPL1971,Aho:SJOC1972,Gries:SOCP1989}.

We show that our algorithm can output a desired partial order of perturbing features for a given ordered action. 
For this purpose, we see that the ordering cost $C_\mathrm{ord}$ of an ordered action only depends on its partially ordered structure obtained by the above procedure.
If $i$ is not an ancestor of $j$ on the DAG of a partially ordered action, then $M_{i,j}=M_{j,i}=0$.
Thus, the actual perturbation in the $k$-th step is calculated as follows:  
\begin{align*} 
    \Delta^{(k)}&(a_{\sigma_1, \dots, \sigma_k} \mid M) \\
    &= a_{\sigma_k} - {\sum}_{l=1}^{k-1} M_{\sigma_l, \sigma_k} \cdot \Delta^{(l)}(a_{\sigma_1, \dots, \sigma_l} \mid M) \\
    &= a_{\sigma_k} - {\sum}_{\substack{l=1,\dots,k-1 \\ \sigma_l \in \operatorname{pa}(\sigma_k)}}  M_{\sigma_l, \sigma_k} \cdot \Delta^{(l)}(a_{\sigma_1, \dots, \sigma_l} \mid M),
\end{align*}
where $\operatorname{pa}(j) \subseteq \supp{a}$ denotes the set of ancestors of a feature $j$ on the DAG of the partially ordered action. 
That is, the ordered action $(a,\sigma')$ has the same ordering cost $C_\mathrm{ord}$ as that of $(a,\sigma)$ if $\sigma'$ is consistent with the partially ordered structure obtained from $(a,\sigma)$.

\section{Experiments}

In this section, we conducted experiments on real datasets to investigate the effectiveness and behavior of our \textbf{OrdCE}.
All the code was implemented in Python 3.7 with scikit-learn and IBM ILOG CPLEX v12.10\footnote{All the code is available at \url{https://github.com/kelicht/ordce}.}. 
All the experiments were conducted on 64-bit macOS Catalina 10.15.6 with Intel Core~i9 2.4GHz CPU and 64GB memory, and we imposed a 300 second time limit for solving.

\subsection{Experimental Setting}
We randomly split each dataset into train (75\%) and test (25\%) instances, and trained $\ell_2$-regularized logistic regression classifiers (LR), random forest classifiers (RF) with $T=100$ decision trees, and two-layer ReLU network classifiers (MLP) with $T=200$ neurons, on each training dataset.
Then, we extracted ordered actions for test instances that had been received undesired prediction results, such as predicted as ``high risk of default" from each classifier. 

\subsubsection{Distance-based Cost Functions}
As a distance-based cost function $C_\mathrm{dist}$, we used four existing cost functions: the total log-percentile shift ({TLPS})~\cite{Ustun:FAT*2019}, weighted $\ell_1$-norm of median absolute deviation ({MAD})~\cite{Wachter:HJLT2018,Russell:FAT*2019}, $\ell_1$-Mahalanobis' distance ({DACE})~\cite{Kanamori:IJCAI2020}, and distance based on structural causal models~(SCM)~\cite{Mahajan:NIPSWS2019}. 
The former two are \emph{norm-based} cost functions that evaluate actions for each perturbing feature independently.
The latter two are \emph{interaction-aware} cost functions that evaluate actions by considering feature-correlation and causality, respectively.

\subsubsection{Baseline Method}
To the best of our knowledge, there is no existing method that determines a minimum-cost perturbing order. 
Even if an ordered action is consistent with a given causal DAG, it is not necessarily optimal with respect to $C_{\mathrm{OrdCE}} = C_{\mathrm{dist}} + \gamma \cdot C_{\mathrm{ord}}$, since $C_{\mathrm{ord}}$ depends not only on the causal direction but also on the amount of the resultant perturbation. 
As a baseline, we proposed a greedy algorithm (\textbf{Greedy}), which consists of the following two steps: 
\begin{enumerate}   
    \item Extract a perturbation vector $a^{\star}$ by optimizing $C_\mathrm{dist}$.
    \item Determine a perturbing order $\sigma$ of $a^\star$ by solving the following optimization problem for $k$ iteratively:
          \begin{align*}
            \sigma_k = \argmin_{d \in \supp{a^{\star}} \setminus \set{\sigma_1, \dots, \sigma_{k-1}}} \left| a^{\star}_d - {\sum}_{l=1}^{k-1} M_{\sigma_l, d} \cdot \Delta^{(l)} \right|.
          \end{align*}
          This procedure greedily selects a perturbing feature that has the smallest cost in each step. 
\end{enumerate}

To compare \textbf{OrdCE} with \textbf{Greedy}, we measured the average values of the distance-based cost $C_\mathrm{dist}$, the ordering cost $C_\mathrm{ord}$, and the objective function $C_\mathrm{OrdCE}$ over the ordered actions obtained by those two methods.

\subsection{Experimental Results}
We show experimental results here.
Owing to page limitation, the detailed results are provided in Appendix~\ref{appendix:add-exp}. 

\subsubsection{Comparison with Baseline}
\begin{table*}[t]
    \centering
    \small
    \subfloat[Objective Function $C_\mathrm{OrdCE}$]{
    \begin{tabular}{cccccccc}
    \toprule
            \multirow{2}{*}{$C_\mathrm{dist}$} & \multirow{2}{*}{Dataset} & \multicolumn{2}{c}{Logistic Regression} & \multicolumn{2}{c}{Random Forest} & \multicolumn{2}{c}{Multilayer Perceptron} \\ \cmidrule(lr){3-4} \cmidrule(lr){5-6} \cmidrule(lr){7-8} 
           & & \textbf{Greedy} & \textbf{OrdCE} & \textbf{Greedy} & \textbf{OrdCE} & \textbf{Greedy} & \textbf{OrdCE}\\
    \midrule
    \multirow{4}{*}{TLPS}     & FICO & 3.96 $\pm$ 2.6 & \textbf{3.27 $\pm$ 2.1} & 3.26 $\pm$ 2.9 & \textbf{3.22 $\pm$ 3.2} & 3.35 $\pm$ 3.0 & \textbf{1.57 $\pm$ 1.4} \\
                              & German & 4.9 $\pm$ 5.8 & \textbf{4.81 $\pm$ 5.7} & 3.23 $\pm$ 2.9 & \textbf{3.2 $\pm$ 2.9} & 5.38 $\pm$ 4.7 & \textbf{5.03 $\pm$ 4.5} \\
                              & WineQuality & 1.78 $\pm$ 1.8 & \textbf{1.57 $\pm$ 1.5} & 0.901 $\pm$ 0.55 & \textbf{0.875 $\pm$ 0.52} & 0.969 $\pm$ 0.83 & \textbf{0.761 $\pm$ 0.61} \\
                              & Diabetes & 2.91 $\pm$ 2.5 & \textbf{2.47 $\pm$ 2.0} & 2.3 $\pm$ 1.8 & \textbf{2.26 $\pm$ 1.8} & 1.12 $\pm$ 1.5 & \textbf{0.668 $\pm$ 0.99} \\
    \midrule
    \multirow{4}{*}{DACE}     & FICO & 10.6 $\pm$ 7.3 & \textbf{9.61 $\pm$ 6.7} & 6.78 $\pm$ 4.8 & \textbf{6.67 $\pm$ 4.7} & 3.5 $\pm$ 3.5 & \textbf{3.41 $\pm$ 3.3} \\
                              & German & 6.19 $\pm$ 5.3 & \textbf{5.88 $\pm$ 4.9} & 5.54 $\pm$ 4.6 & \textbf{5.42 $\pm$ 4.5} & 7.0 $\pm$ 5.8 & \textbf{6.7 $\pm$ 5.4} \\
                              & WineQuality & 2.93 $\pm$ 2.0 & \textbf{2.42 $\pm$ 1.6} & 1.65 $\pm$ 1.2 & \textbf{1.51 $\pm$ 1.1} & 1.91 $\pm$ 1.5 & \textbf{1.66 $\pm$ 1.3} \\
                              & Diabetes & 2.56 $\pm$ 1.7 & \textbf{2.43 $\pm$ 1.6} & 2.38 $\pm$ 1.7 & \textbf{2.21 $\pm$ 1.6} & 0.832 $\pm$ 1.2 & \textbf{0.766 $\pm$ 1.1} \\
    \bottomrule
    \end{tabular}
    }
    \hfill
    \subfloat[Ordering Cost Function $C_\mathrm{ord}$]{
    \begin{tabular}{cccccccc}
    \toprule
            \multirow{2}{*}{$C_\mathrm{dist}$} & \multirow{2}{*}{Dataset} & \multicolumn{2}{c}{Logistic Regression} & \multicolumn{2}{c}{Random Forest} & \multicolumn{2}{c}{Multilayer Perceptron} \\ \cmidrule(lr){3-4} \cmidrule(lr){5-6} \cmidrule(lr){7-8} 
           & & \textbf{Greedy} & \textbf{OrdCE} & \textbf{Greedy} & \textbf{OrdCE} & \textbf{Greedy} & \textbf{OrdCE}\\
    \midrule
    \multirow{4}{*}{TLPS}     & FICO & 2.21 $\pm$ 1.6 & \textbf{1.33 $\pm$ 0.87} & 1.72 $\pm$ 1.5 & \textbf{1.49 $\pm$ 1.2} & 3.05 $\pm$ 2.8 & \textbf{0.84 $\pm$ 0.74} \\
                              & German & 1.85 $\pm$ 1.5 & \textbf{1.71 $\pm$ 1.4} & 1.5 $\pm$ 1.2 & \textbf{1.47 $\pm$ 1.2} & 2.27 $\pm$ 1.8 & \textbf{1.76 $\pm$ 1.6} \\
                              & WineQuality & 1.0 $\pm$ 0.98 & \textbf{0.765 $\pm$ 0.59} & 0.475 $\pm$ 0.31 & \textbf{0.439 $\pm$ 0.29} & 0.69 $\pm$ 0.63 & \textbf{0.446 $\pm$ 0.35} \\
                              & Diabetes & 1.74 $\pm$ 1.6 & \textbf{1.01 $\pm$ 0.81} & 0.939 $\pm$ 0.67 & \textbf{0.883 $\pm$ 0.64} & 0.862 $\pm$ 1.3 & \textbf{0.318 $\pm$ 0.58} \\
    \midrule
    \multirow{4}{*}{DACE}     & FICO & 3.79 $\pm$ 2.6 & \textbf{2.41 $\pm$ 1.8} & 2.14 $\pm$ 1.5 & \textbf{1.59 $\pm$ 1.2} & 1.24 $\pm$ 1.2 & \textbf{0.918 $\pm$ 0.86} \\
                              & German & 1.92 $\pm$ 1.8 & \textbf{1.43 $\pm$ 1.1} & 1.6 $\pm$ 1.3 & \textbf{1.46 $\pm$ 1.2} & 2.23 $\pm$ 2.0 & \textbf{1.87 $\pm$ 1.5} \\
                              & WineQuality & 1.31 $\pm$ 0.94 & \textbf{0.781 $\pm$ 0.53} & 0.716 $\pm$ 0.54 & \textbf{0.503 $\pm$ 0.34} & 0.796 $\pm$ 0.69 & \textbf{0.509 $\pm$ 0.41} \\
                              & Diabetes & 1.2 $\pm$ 0.83 & \textbf{1.02 $\pm$ 0.7} & 1.13 $\pm$ 0.84 & \textbf{0.912 $\pm$ 0.67} & 0.425 $\pm$ 0.63 & \textbf{0.322 $\pm$ 0.47} \\
    \bottomrule
    \end{tabular}
    }
    \caption{Experimental results on the real datasets.}
    \label{tab:real:comp}
\end{table*}

We used four real datasets: FICO ($D=23$)~\cite{fico:2018}, German ($D=40$), WineQuality ($D=12$), and Diabetes ($D=8$)~\cite{Dua:2019} datasets, where $D$ is the number of features.
For German dataset, we transformed each categorical feature into a one-hot encoded vector.
For each dataset, we estimated the adjacency matrix of a causal DAG by the DirectLiNGAM algorithm~\cite{Shimizu:JMLR2011,Hyvarinen:JMLR2013}, and computed an interaction matrix $M$ from the adjacency matrix (see Section~3). 
We set $\gamma=1.0$ and $K=4$.
Here we show the reults of TLPS and DACE. 

\cref{tab:real:comp}(a) and \cref{tab:real:comp}(b) present the average of the objective function $C_\mathrm{OrdCE}$ and the ordering cost $C_\mathrm{ord}$ for extracted ordered actions, respectively.
From \cref{tab:real:comp}, we can observe that \textbf{OrdCE} always achieved lower $C_\mathrm{OrdCE}$ and $C_\mathrm{ord}$ than \textbf{Greedy} for all datasets and classifiers.
Especially, in MLP and TLPS on FICO dataset, the averages of $C_{\mathrm{OrdCE}}$ and $C_{\mathrm{dist}}$ given by \textbf{OrdCE} are 1.57 and 0.84, respectively, which were less than half of those obtained by \textbf{Greedy}.

\begin{table*}[t]
    \centering
    \small
    \subfloat[TLPS~\protect\cite{Ustun:FAT*2019}]{
    \begin{tabular}{lclccc}
        \toprule
        \textbf{Method} & \textbf{Order} & \textbf{Feature} & \textbf{Action} & $C_\mathrm{dist}$ & $C_\mathrm{ord}$ \\ 
        \toprule
        {\textbf{\textbf{Greedy}}}   & 1st & {``BMI"}     & -6.25  & {\textbf{0.778}} & {0.828} \\
        \midrule
        \multirow{2}{*}{\textbf{OrdCE}} & 1st & {``Glucose"} & -3.0  & \multirow{2}{*}{0.825} & \multirow{2}{*}{\textbf{0.749}} \\
                                        & 2nd & {``BMI"}     & -5.05 & & \\
        \bottomrule
    \end{tabular}
    }
    \hfill
    \subfloat[DACE~\protect\cite{Kanamori:IJCAI2020}]{
    \begin{tabular}{lclccc}
        \toprule
        \textbf{Method} & \textbf{Order} & \textbf{Feature} & \textbf{Action} & $C_\mathrm{dist}$ & $C_\mathrm{ord}$ \\ 
        \toprule
        \multirow{4}{*}{\textbf{Greedy}}         & 1st & {``BMI"}           & -0.8  & \multirow{4}{*}{\textbf{0.716}} & \multirow{4}{*}{0.825} \\
                                        & 2nd & {``SkinThickness"} & -2.5  & & \\
                                        & 3rd & {``Glucose"}       & -8.5  & & \\
                                        & 4th & {``Insulin"}       & -32.0 & & \\
        \midrule
        \multirow{4}{*}{\textbf{OrdCE}} & 1st & {``Insulin"}       & -32.0 & \multirow{4}{*}{\textbf{0.716}} & \multirow{4}{*}{\textbf{0.528}} \\
                                        & 2nd & {``Glucose"}       & -8.5  & & \\
                                        & 3rd & {``SkinThickness"} & -2.5  & & \\
                                        & 4th & {``BMI"}           & -0.8  & & \\
        \bottomrule
    \end{tabular}
    }

    \caption{Examples of ordered actions extracted from the RF classifier on the Diabetes dataset.
             }
    \label{tab:real:example}
\end{table*}
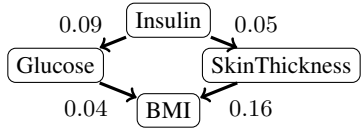
\begin{figure}[t]
    \centering
    \begin{tikzpicture}
        \node[draw, rectangle,rounded corners=3pt] (a) at (0,0.6) {\small Insulin};
        \node[draw, rectangle,rounded corners=3pt] (b) at (-1.5,0) {\small Glucose};
        \node[draw, rectangle,rounded corners=3pt] (c) at (1.5,0) {\small SkinThickness};
        \node[draw, rectangle,rounded corners=3pt] (d) at (0,-0.6) {\small BMI};
        \draw[very thick, ->] (a)--(b) node [midway, above left] {\small $0.09$};
        \draw[very thick, ->] (a)--(c) node [midway, above right] {\small $0.05$};
        \draw[very thick, ->] (b)--(d) node [midway, below left] {\small $0.04$};
        \draw[very thick, ->] (c)--(d) node [midway, below right] {\small $0.16$};
    \end{tikzpicture}
    \caption{Subgraph of the causal DAG of the Diabetes dataset estimated by the DirectLiNGAM algorithm~\cite{Shimizu:JMLR2011,Hyvarinen:JMLR2013}.}
    \label{fig:real:causaldag}
\end{figure}

Next, we examine the ordered actions given by \textbf{OrdCE} to confirm the practicality. 
\cref{tab:real:example} presents examples of ordered actions extracted from the RF classifier on the Diabetes dataset, and \cref{fig:real:causaldag} presents a subgraph of the estimated causal DAG of the dataset. 
In both cases of TLPS and DACE, our \textbf{OrdCE} output ordered actions that accords with the directed edges in the causal DAG in \cref{fig:real:causaldag}. 
On the other hand, the action extracted by \textbf{Greedy} with DACE, the order is not consistent with the causal DAG.
From these results, we confirmed that \textbf{OrdCE} succeeded in obtaining a reasonable perturbing order from the perspective of the causal DAG.
In addition, for TLPS, the perturbation given by \textbf{OrdCE} is different from that given by \textbf{Greedy}.
This difference is caused by the effect that \textbf{OrdCE} optimizes a perturbation vector and its order simultaneously. 

Regarding the computation time, \textbf{OrdCE} was certainly slower than \textbf{Greedy} because \textbf{OrdCE} exactly solved Problem~\ref{prob:ordce}. 
In MLP and TLPS on FICO dataset, the average computation times of \textbf{OrdCE} and \textbf{Greedy} are 183 and 12.6 seconds, respectively. 
For other datasets, \textbf{OrdCE} is within 1.38--120 times slower than \textbf{Greedy}. 
However, \cref{tab:real:comp,tab:real:example} indicate that \textbf{OrdCE} found better ordered actions in terms of $C_\mathrm{OrdCE}$ and $C_\mathrm{ord}$ than \textbf{Greedy} within 300 seconds, which is a reasonable computation time. 
The results of computation time are presented in Appendix~\ref{appendix:add-exp}.

\subsubsection{Sensitivity Analysis of Trade-off Parameter}
\begin{figure}[t]
    \centering
    \includegraphics[width=0.7\columnwidth]{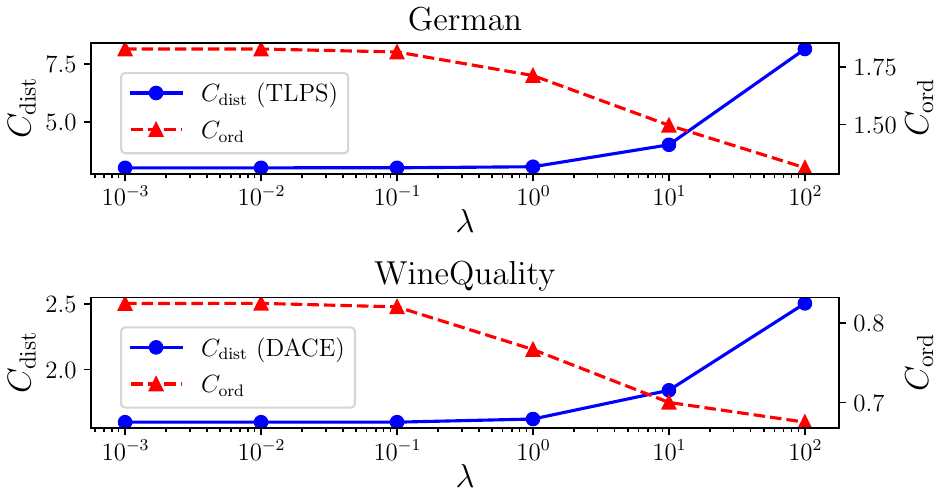}
    \caption{Sensitivity analyses of the trade-off parameter $\gamma$ of \textbf{OrdCE} between the average $C_\mathrm{dist}$ and $C_\mathrm{ord}$.}
    \label{fig:real:sens}
\end{figure}

To examine the sensitivity of the trade-off parameter $\gamma$, we observed $C_\mathrm{dist}$ and $C_\mathrm{ord}$ of ordered actions extracted from LR classifiers by varying $\gamma$.
The above (resp.\ below) figure in \cref{fig:real:sens} presents the average $C_\mathrm{dist}$ (resp.\ $C_\mathrm{ord}$) for each $\gamma$ on the German and WineQuality datasets. 
We can see trade-off relationship between $C_\mathrm{dist}$ and $C_\mathrm{ord}$. 
As mentioned in~\cite{Wachter:HJLT2018,Mothilal:FAT*2020}, suggesting multiple actions is helpful for diversity. 
By varying $\gamma$, we can obtain several distinct ordered actions that have diverse characteristics in terms of $C_\mathrm{dist}$ and $C_\mathrm{ord}$.

\section{Conclusion}
We proposed Ordered Counterfactual Explanation (OrdCE) that provides an optimal pair of a perturbation vector and an order of the features to be perturbed.
We introduced a new objective function that evaluates the required cost of a perturbation vector and an order, and proposed a MILO formulation for optimizing it.
By experiments on real datasets, we confirmed the effectiveness of our method by comparing it with a greedy method.
As future work, we plan to conduct user-experiments to evaluate the usability of our OrdCE.
In this study, while we assume the causal relationship is linear, our cost function has a potential to deal with non-linear relationships, which sometimes appear in the real world~\cite{Pearl:2009}.
Therefore, it is also interesting future work to develop a method for optimizing our cost function with non-linear causal relationships.

\section*{Acknowledgments}
We wish to thank Kunihiro Wasa and Kazuhiro Kurita for making a number of valuable suggestions. 
We also thank anonymous reviewers for their insightful comments. 
This work was supported in part by JSPS KAKENHI Grant-in-Aid for JSPS Research Fellow 20J20654, Scientific Research (A) 20H00595, and JST CREST JPMJCR18K3.

\appendix

\section{Additional Experimental Results}\label{appendix:add-exp}

\subsection{Comparison with Baseline}
\begin{table*}[t]
    \centering
    {\small
    \subfloat[Objective Function $C_\mathrm{OrdCE}$]{
    \begin{tabular}{cccccccc}
    \toprule
            \multirow{2}{*}{$C_\mathrm{dist}$} & \multirow{2}{*}{Dataset} & \multicolumn{2}{c}{Logistic Regression} & \multicolumn{2}{c}{Random Forest} & \multicolumn{2}{c}{Multilayer Perceptron} \\ \cmidrule(lr){3-4} \cmidrule(lr){5-6} \cmidrule(lr){7-8} 
           & & \textbf{Greedy} & \textbf{OrdCE} & \textbf{Greedy} & \textbf{OrdCE} & \textbf{Greedy} & \textbf{OrdCE}\\
    \midrule
    \multirow{4}{*}{MAD}     & FICO & 3.19 $\pm$ 2.1 & \textbf{3.07 $\pm$ 2.0} & 2.94 $\pm$ 2.7 & \textbf{2.85 $\pm$ 1.9} & 1.09 $\pm$ 0.97 & \textbf{1.07 $\pm$ 0.98} \\
                              & German & 5.90 $\pm$ 5.0 & \textbf{2.52 $\pm$ 1.6} & 6.70 $\pm$ 8.8 & \textbf{1.94 $\pm$ 1.6} & 6.53 $\pm$ 6.8 & \textbf{5.14 $\pm$ 4.2} \\
                              & Wine & 1.44 $\pm$ 1.0 & \textbf{1.43 $\pm$ 1.0} & 0.982 $\pm$ 0.59 & \textbf{0.966 $\pm$ 0.58} & 0.872 $\pm$ 0.68 & \textbf{0.865 $\pm$ 0.68} \\
                              & Diabetes & \textbf{1.82 $\pm$ 1.2} & \textbf{1.82 $\pm$ 1.2} & 1.34 $\pm$ 1.2 & \textbf{1.31 $\pm$ 1.1} & 0.695 $\pm$ 0.71 & \textbf{0.683 $\pm$ 0.70} \\
    \midrule
    \multirow{4}{*}{SCM}     & FICO & 3.96 $\pm$ 3.1 & \textbf{3.41 $\pm$ 2.5} & 3.80 $\pm$ 2.9 & \textbf{3.53 $\pm$ 2.8} & 2.23 $\pm$ 2.0 & \textbf{2.09 $\pm$ 1.8} \\
                              & German & 3.60 $\pm$ 3.2 & \textbf{2.95 $\pm$ 2.4} & 2.05 $\pm$ 2.0 & \textbf{1.73 $\pm$ 1.6} & 3.99 $\pm$ 3.2 & \textbf{3.42 $\pm$ 2.8} \\
                              & Wine & 2.32 $\pm$ 1.6 & \textbf{1.75 $\pm$ 1.2} & 1.43 $\pm$ 0.95 & \textbf{1.32 $\pm$ 0.90} & 1.21 $\pm$ 0.9 & \textbf{1.03 $\pm$ 0.75} \\
                              & Diabetes & \textbf{1.78 $\pm$ 1.2} & \textbf{1.78 $\pm$ 1.2} & 1.34 $\pm$ 1.2 & \textbf{1.21 $\pm$ 1.0} & 0.629 $\pm$ 0.87 & \textbf{0.549 $\pm$ 0.74} \\
    \bottomrule
    \end{tabular}
    }
    \hfill
    \subfloat[Ordering Cost Function $C_\mathrm{ord}$]{
    \begin{tabular}{cccccccc}
    \toprule
            \multirow{2}{*}{$C_\mathrm{dist}$} & \multirow{2}{*}{Dataset} & \multicolumn{2}{c}{Logistic Regression} & \multicolumn{2}{c}{Random Forest} & \multicolumn{2}{c}{Multilayer Perceptron} \\ \cmidrule(lr){3-4} \cmidrule(lr){5-6} \cmidrule(lr){7-8} 
           & & \textbf{Greedy} & \textbf{OrdCE} & \textbf{Greedy} & \textbf{OrdCE} & \textbf{Greedy} & \textbf{OrdCE}\\
    \midrule
    \multirow{4}{*}{MAD}      & FICO & 1.41 $\pm$ 0.91 & \textbf{1.28 $\pm$ 0.80} & 1.45 $\pm$ 1.1 & \textbf{1.37 $\pm$ 0.84} & 0.562 $\pm$ 0.50 & \textbf{0.526 $\pm$ 0.50} \\
                              & German & 5.33 $\pm$ 4.8 & \textbf{1.52 $\pm$ 1.1} & 5.99 $\pm$ 8.5 & \textbf{0.975 $\pm$ 1.2} & 3.48 $\pm$ 4.7 & \textbf{1.64 $\pm$ 1.3} \\
                              & Wine & 0.757 $\pm$ 0.54 & \textbf{0.750 $\pm$ 0.54} & 0.497 $\pm$ 0.32 & \textbf{0.470 $\pm$ 0.30} & 0.449 $\pm$ 0.35 & \textbf{0.439 $\pm$ 0.35} \\
                              & Diabetes & \textbf{0.890 $\pm$ 0.60} & \textbf{0.890 $\pm$ 0.60} & 0.636 $\pm$ 0.55 & \textbf{0.588 $\pm$ 0.51} & 0.303 $\pm$ 0.4 & \textbf{0.283 $\pm$ 0.39} \\
    \midrule
    \multirow{4}{*}{SCM}      & FICO & 1.98 $\pm$ 1.6 & \textbf{1.32 $\pm$ 0.84} & 1.90 $\pm$ 1.5 & \textbf{1.49 $\pm$ 1.0} & 1.08 $\pm$ 1.0 & \textbf{0.813 $\pm$ 0.73} \\
                              & German & 2.02 $\pm$ 1.9 & \textbf{1.35 $\pm$ 1.1} & 1.11 $\pm$ 1.2 & \textbf{0.796 $\pm$ 0.73} & 2.16 $\pm$ 1.8 & \textbf{1.59 $\pm$ 1.3} \\
                              & Wine & 1.34 $\pm$ 0.95 & \textbf{0.769 $\pm$ 0.54} & 0.802 $\pm$ 0.57 & \textbf{0.583 $\pm$ 0.41} & 0.645 $\pm$ 0.5 & \textbf{0.455 $\pm$ 0.35} \\
                              & Diabetes & 0.890 $\pm$ 0.60 & \textbf{0.889 $\pm$ 0.60} & 0.716 $\pm$ 0.66 & \textbf{0.577 $\pm$ 0.50} & 0.349 $\pm$ 0.5 & \textbf{0.264 $\pm$ 0.36} \\
    \bottomrule
    \end{tabular}
    }    
    }
    \caption{Experimental results on the real datasets.}
    \label{tab:real:appendix:comp}
\end{table*}

\begin{table*}[t]
    \centering
    {\small
    \begin{tabular}{cccccccc}
    \toprule
            \multirow{2}{*}{$C_\mathrm{dist}$} & \multirow{2}{*}{Dataset} & \multicolumn{2}{c}{Logistic Regression} & \multicolumn{2}{c}{Random Forest} & \multicolumn{2}{c}{Multilayer Perceptron} \\ \cmidrule(lr){3-4} \cmidrule(lr){5-6} \cmidrule(lr){7-8} 
           & & \textbf{Greedy} & \textbf{OrdCE} & \textbf{Greedy} & \textbf{OrdCE} & \textbf{Greedy} & \textbf{OrdCE}\\
    \midrule
    \multirow{4}{*}{TLPS}     & FICO & 0.110 $\pm$ 0.023 & 12.3 $\pm$ 15 & 21.1 $\pm$ 50 & 125 $\pm$ 130 & 12.6 $\pm$ 48 & 183 $\pm$ 140 \\
                              & German & 0.0226 $\pm$ 0.0052 & 0.528 $\pm$ 0.57 & 10.6 $\pm$ 3.8 & 15.6 $\pm$ 8.8 & 0.154 $\pm$ 0.08 & 1.44 $\pm$ 0.91 \\
                              & Wine & 0.0682 $\pm$ 0.013 & 2.32 $\pm$ 4.5 & 8.82 $\pm$ 3.8 & 25.9 $\pm$ 48 & 0.354 $\pm$ 0.16 & 4.88 $\pm$ 13 \\
                              & Diabetes & 0.0339 $\pm$ 0.0046 & 0.720 $\pm$ 0.52 & 6.18 $\pm$ 7.4 & 42.4 $\pm$ 60 & 0.290 $\pm$ 0.1 & 4.81 $\pm$ 12 \\
    \midrule
    \multirow{4}{*}{DACE}     & FICO & 21.6 $\pm$ 22 & 277 $\pm$ 75 & 146 $\pm$ 78 & 287 $\pm$ 48 & 190 $\pm$ 140 & 233 $\pm$ 120 \\
                              & German & 0.260 $\pm$ 0.15 & 3.33 $\pm$ 5.3 & 30.2 $\pm$ 15 & 138 $\pm$ 180 & 0.778 $\pm$ 0.37 & 6.25 $\pm$ 9.8 \\
                              & Wine & 2.04 $\pm$ 0.81 & 150 $\pm$ 120 & 137 $\pm$ 110 & 254 $\pm$ 89 & 11.9 $\pm$ 21 & 121 $\pm$ 120 \\
                              & Diabetes & 0.245 $\pm$ 0.077 & 22.8 $\pm$ 52 & 25.4 $\pm$ 26 & 217 $\pm$ 120 & 2.27 $\pm$ 3.1 & 50.2 $\pm$ 100 \\
    \midrule
    \multirow{4}{*}{MAD}      & FICO & 0.0981 $\pm$ 0.021 & 8.05 $\pm$ 9.1 & 18.3 $\pm$ 27 & 117 $\pm$ 120 & 13.3 $\pm$ 32 & 151 $\pm$ 140 \\
                              & German & 0.0169 $\pm$ 0.0034 & 0.579 $\pm$ 0.54 & 8.09 $\pm$ 1.3 & 81.2 $\pm$ 110 & 0.0892 $\pm$ 0.058 & 5.15 $\pm$ 7.6 \\
                              & Wine & 0.0764 $\pm$ 0.019 & 1.16 $\pm$ 0.49 & 12.6 $\pm$ 5.5 & 40.2 $\pm$ 68 & 0.319 $\pm$ 0.12 & 8.13 $\pm$ 29 \\
                              & Diabetes & 0.0274 $\pm$ 0.0041 & 0.449 $\pm$ 0.13 & 8.03 $\pm$ 8.2 & 37.4 $\pm$ 61 & 0.253 $\pm$ 0.083 & 3.77 $\pm$ 8.3 \\
    \midrule
    \multirow{4}{*}{SCM}      & FICO & 0.564 $\pm$ 0.35 & 212 $\pm$ 120 & 96.1 $\pm$ 80 & 275 $\pm$ 70 & 175 $\pm$ 140 & 231 $\pm$ 120 \\
                              & German & 0.0360 $\pm$ 0.0085 & 0.829 $\pm$ 0.73 & 13.2 $\pm$ 8.7 & 63.4 $\pm$ 48 & 0.133 $\pm$ 0.09 & 1.61 $\pm$ 1.6 \\
                              & Wine & 0.475 $\pm$ 0.19 & 106 $\pm$ 110 & 123 $\pm$ 100 & 263 $\pm$ 82 & 1.92 $\pm$ 1.2 & 91.4 $\pm$ 100 \\
                              & Diabetes & 0.0450 $\pm$ 0.021 & 2.10 $\pm$ 5.1 & 18.1 $\pm$ 14 & 143 $\pm$ 120 & 0.806 $\pm$ 0.53 & 42.4 $\pm$ 98 \\
    \bottomrule
    \end{tabular}
    }
    \caption{Computational times on the real datasets.}
    \label{tab:real:appendix:time}
\end{table*}

\cref{tab:real:appendix:comp}(a) and \cref{tab:real:appendix:comp}(b) present the average values of the objective function $C_\mathrm{OrdCE}$ and the ordering cost $C_\mathrm{ord}$ for the distance-based cost functions MAD and SCM, respectively.
The result of computation time for each distance-based cost function, i.e., TLPS, DACE, MAD, and SCM, is shown in \cref{tab:real:appendix:time}.

\subsection{Sensitivity Analysis of Trade-off Parameter}
\begin{figure*}
    \centering
    \subfloat[TLPS]{
        \includegraphics[width=0.4\linewidth]{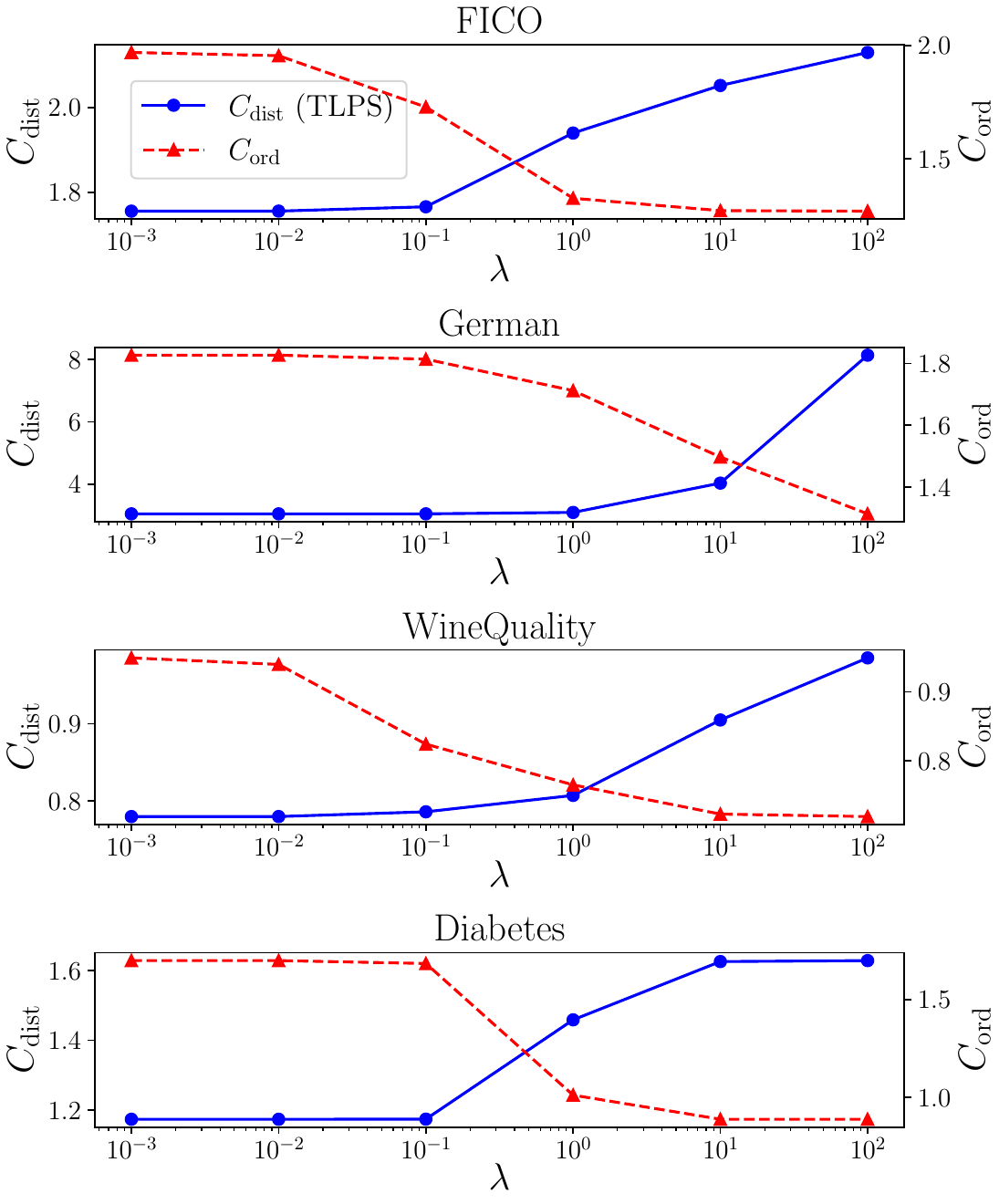}
    }
    \hspace{10mm}
    \subfloat[DACE]{
        \includegraphics[width=0.4\linewidth]{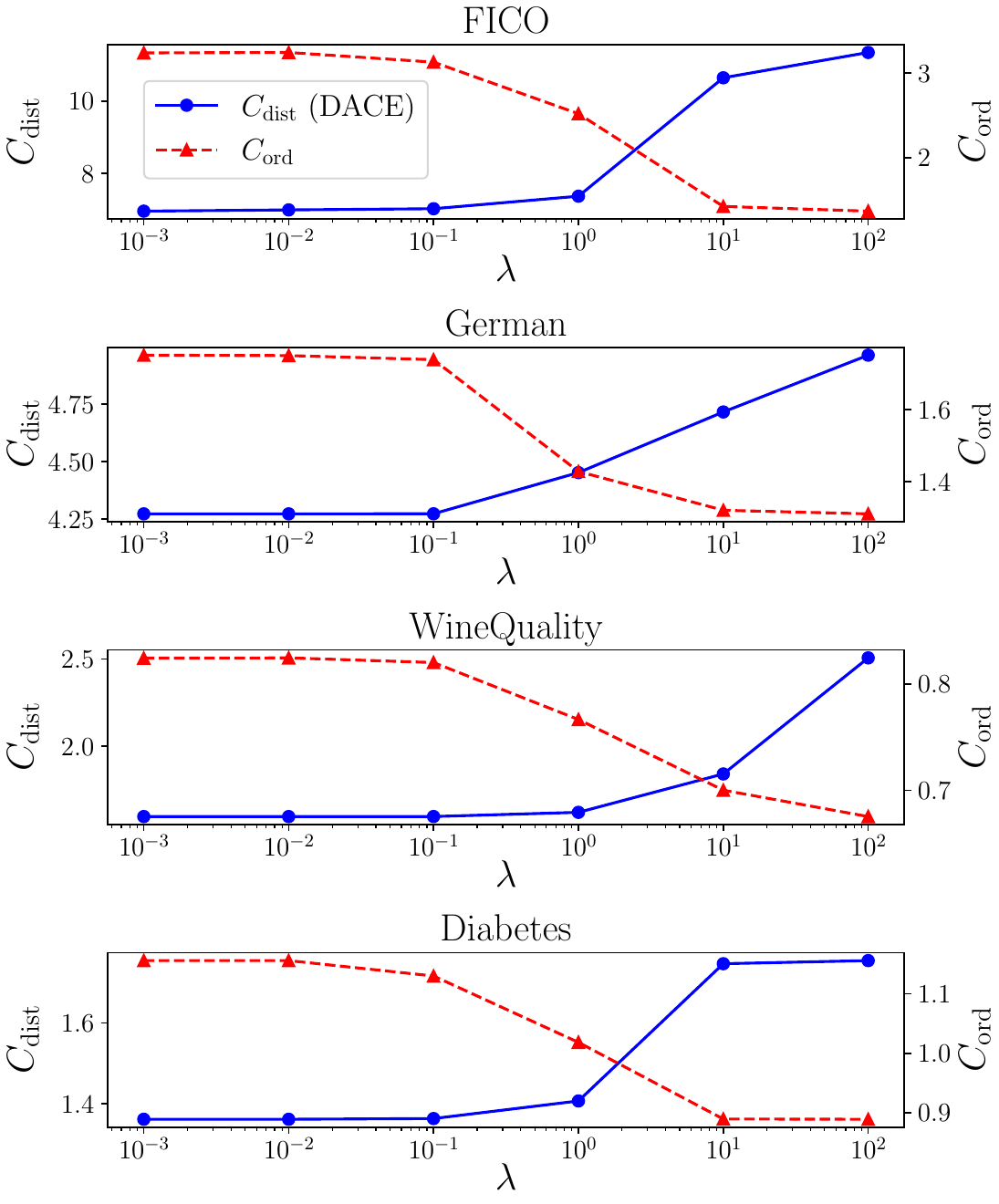}
    }
    \hfill
    \subfloat[MAD]{
        \includegraphics[width=0.4\linewidth]{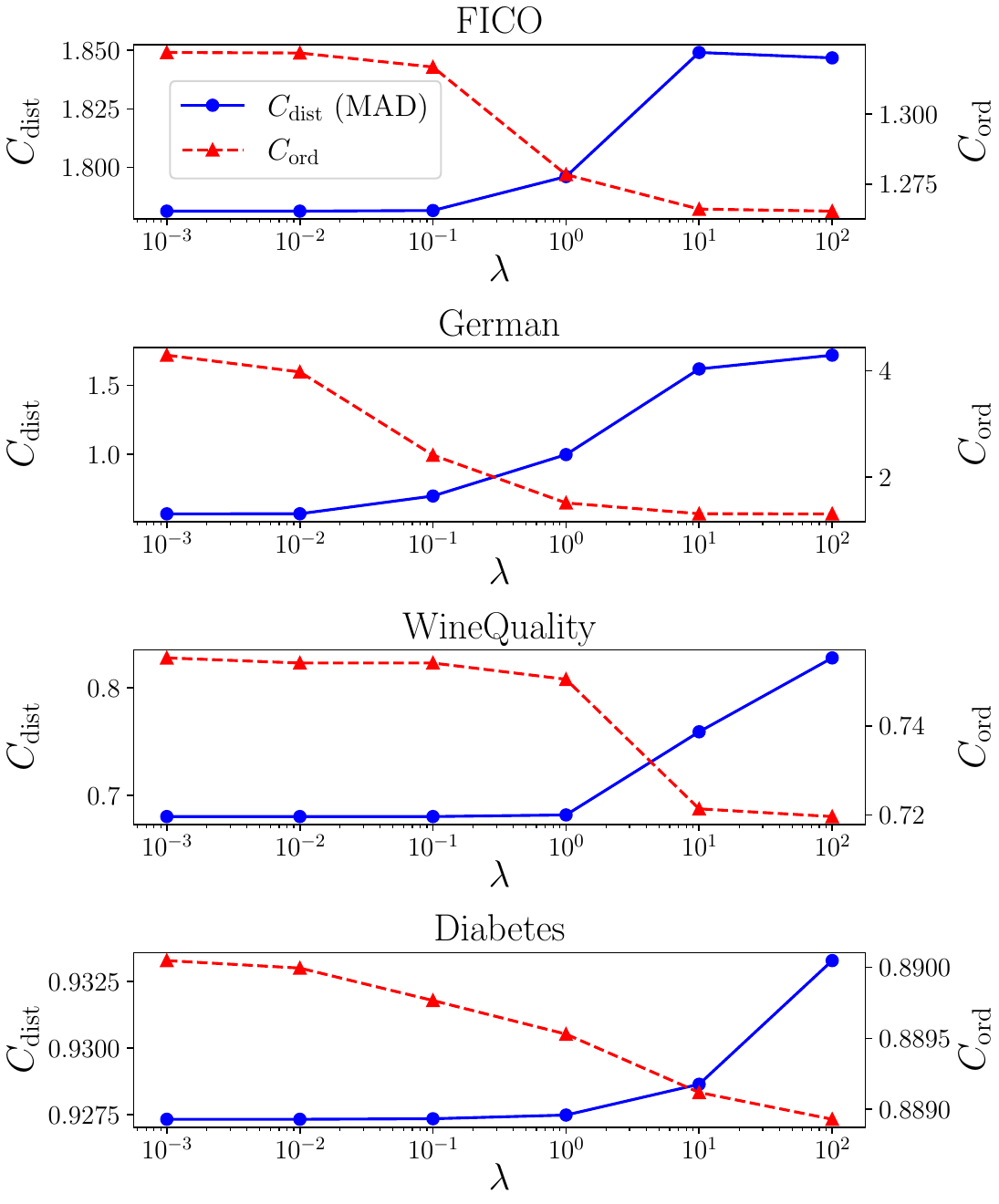}
    }
    \hspace{10mm}
    \subfloat[SCM]{
        \includegraphics[width=0.4\linewidth]{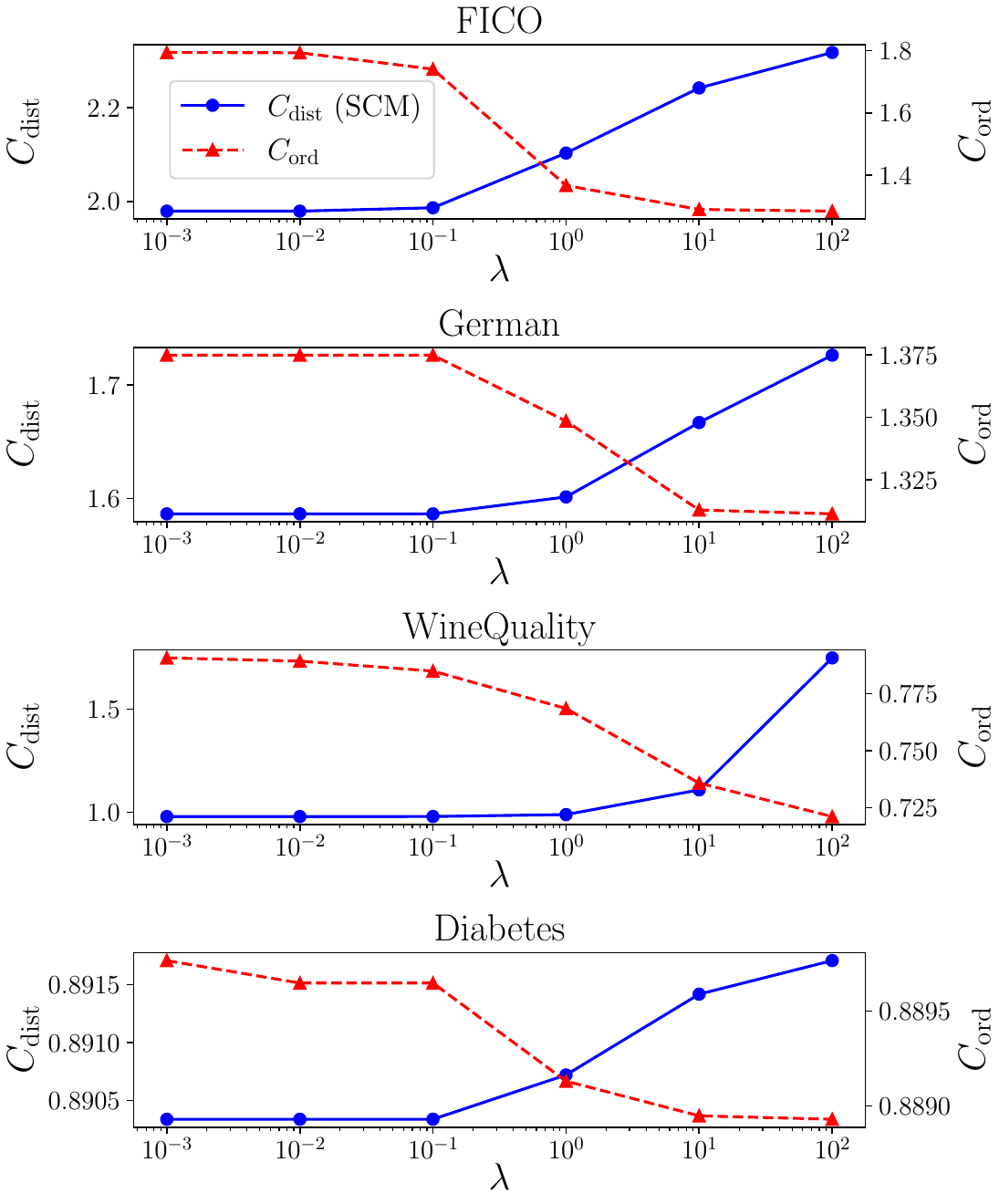}
    }
    \caption{Sensitivity analyses of the trade-off parameter $\gamma$ of \textbf{OrdCE} between the average $C_\mathrm{dist}$ and $C_\mathrm{ord}$.}
    \label{fig:real:appendix:sens}
\end{figure*}

The result of sensitivity analyses of the trade-off parameter~$\gamma$ for each distance-based cost function and real dataset is shown in \cref{fig:real:appendix:sens}.

\clearpage

\bibliographystyle{plain}
\bibliography{ref}

\end{document}